\documentclass{article}
\pdfoutput=1  

\usepackage[preprint, nonatbib]{neurips_2023}

\usepackage[utf8]{inputenc} 
\usepackage[T1]{fontenc}    
\usepackage{url}            
\usepackage{booktabs}       
\usepackage{amsfonts}       
\usepackage{nicefrac}       
\usepackage{microtype}      
\usepackage{xcolor}         

\usepackage[colorlinks, linkcolor=red, anchorcolor=red, citecolor=blue, breaklinks]{hyperref}

\usepackage{microtype}
\usepackage{graphicx}
\usepackage{subfigure}
\usepackage{booktabs} 

\usepackage{amsmath}
\usepackage{amssymb}
\usepackage{mathtools}
\usepackage{amsthm}

\usepackage{listings}
\usepackage{multirow}
\usepackage{float}

\usepackage{xcolor}
\usepackage{wrapfig}
\usepackage{colortbl}
\usepackage[makeroom]{cancel}
\usepackage{cases}

\definecolor{gray}{rgb}{0.5,0.5,0.5}
\definecolor{darkergreen}{RGB}{21, 152, 56}
\definecolor{RoyalBlue}{RGB}{65,105,225}
\definecolor{YellowOrange}{RGB}{255,165,0}
\definecolor{gray94}{gray}{.94}
\definecolor{gray90}{gray}{.90}

\newcommand{\blue}[1]{\textcolor{blue}{#1}}

\title{Boosting Discriminative Visual Representation Learning with Scenario-Agnostic Mixup}

\author{
    Siyuan Li\thanks{Equal contribution.\ \ \ $^\dag$Corrsponding Author.}\ \ \ \ \ \ 
    Zicheng Liu$^*$\ \ \ \ 
    Zedong Wang$^*$\ \ \ \ 
    Di Wu\ \ \ \ 
    Zihan Liu\ \ \ \ 
    Stan Z. Li$^\dag$\\
    AI Lab, Research Center for Industries of the Future, Westlake University, Hangzhou, China\\
    \{lisiyuan,~liuzicheng,~wangzedong,~wudi,~liuzihan,~stan.zq.li\}@westlake.edu.cn
    \vspace{-1em} 
}

\begin{document}

\maketitle

\begin{abstract}

Mixup is a well-known data-dependent augmentation technique for DNNs, consisting of two sub-tasks: mixup generation and classification. 
However, the recent dominant online training method confines mixup to supervised learning (SL), and the objective of the generation sub-task is limited to selected sample pairs instead of the whole data manifold, which might cause trivial solutions.
To overcome such limitations, we comprehensively study the objective of mixup generation and propose \textbf{S}cenario-\textbf{A}gnostic \textbf{Mix}up  (SAMix) for both SL and Self-supervised Learning (SSL) scenarios. 
Specifically, we hypothesize and verify the objective function of mixup generation as optimizing local smoothness between two mixed classes subject to global discrimination from other classes.
Accordingly, we propose $\eta$-balanced mixup loss for complementary learning of the two sub-objectives. 
Meanwhile, a label-free generation sub-network is designed, which effectively provides non-trivial mixup samples and improves transferable abilities. 
Moreover, to reduce the computational cost of online training, we further introduce a pre-trained version, SAMix$^\mathcal{P}$, achieving more favorable efficiency and generalizability.
Extensive experiments on nine SL and SSL benchmarks demonstrate the consistent superiority and versatility of SAMix compared with existing methods.
%

\end{abstract}
\section{Introduction}
\label{sec:intro}

\begin{wrapfigure}{r}{0.5\linewidth}
  \vspace{-5.0em}
  \begin{center}
    \includegraphics[width=0.95\linewidth]{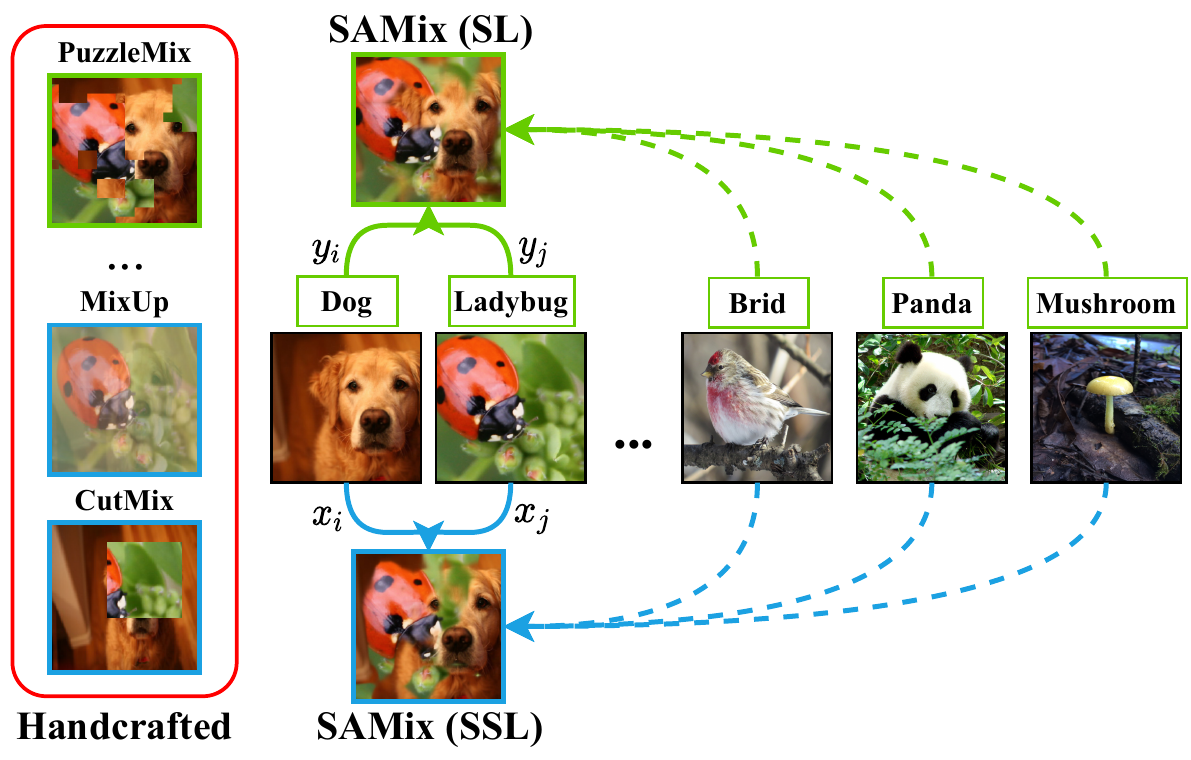}
  \end{center}
  \vspace{-1.5em}
  \caption{
    Illustration of SAMix. Labels are required for the mixed samples in green boxes but not for the blue ones.
    The solid line indicates that the local relationship directly influences the mixed sample, whereas the dashed one denotes that samples of the other classes serve as global constraints on the current mixed data. 
    Compared to handcrafted policies focusing only on local sample pairs, SAMix can exploit local and global information for mixup generation while exerting no dependency on labels.
  }
  \label{fig:intro}
  \vspace{-2.0em}
\end{wrapfigure}

One of the fundamental problems in machine learning is how to learn a proper low-dimensional representation that captures intrinsic data structures and facilitates downstream tasks in an efficient manner~\cite{bengio2013representation, devlin2018bert, korbar2018cooperative, icml2020simclr}. 
Data mixing, as a means of generating symmetric mixed data and labels, has largely improved the efficiency of deep neural networks (DNNs) in learning discriminative representation across various scenarios~\cite{zhang2017mixup, kim2020puzzle, 2021iclrimix}, especially for Vision Transformers (ViTs)~\cite{iclr2021vit, icml2021DeiT}. 
Despite its general application, the policy of sample generation process in data mixing requires an \textit{explicit} design (\textit{e.g.} linear interpolation or random local patch replacement).
%
Differ from that, the optimizable data mixing algorithms utilize labels to localize task-relevant targets (\textit{e.g.}, gradCAM~\cite{selvaraju2017gradcam}) and thereby generate semantic mixed samples such as offline maximizing saliency information of related samples~\cite{kim2020puzzle, kim2021comixup, uddin2020saliencymix}. 
These handcrafted mixing policies are shown in the red box of Figure~\ref{fig:intro}. 
Additionally, \cite{eccv2020automix} learns a mixup generator by supervised adversarial training.

Most current works combine mixup with contrastive learning, which directly transfers linear mixup methods into contrastive learning~\cite{2021iclrimix, nips2020mochi, 2022aaaiunmix}. 
The recently proposed AutoMix~\cite{liu2022automix} provides a novel perspective to make mixup policy parameterized and can be trained online.
Although these optimizable methods have attained significant gains in supervised learning (SL) tasks, they still do not exploit the underlying structure of the whole observed data manifold, resulting in trivial solutions without the guidance of labels, which makes them fail to apply in self-supervised learning (SSL) scenarios (discussed in Section~\ref{sec:problem}).
The question then naturally arises as to whether it is possible to design a more generalized and trained mixup policy that can be applied to both SL and SSL scenarios.
To achieve this goal, there are two remaining open challenges to be solved: 
\textbf{
(i) how to solve the problem of trivial solutions in online training approaches; 
(ii) how to design a proper generation objective to make the algorithm generalizable for both SL and SSL.
}

In this paper, we propose \textbf{S}cenario-\textbf{A}gnostic \textbf{Mix}up (SAMix), a framework shown in Figure \ref{fig:intro} that employs \textit{$\eta$-balanced mixup loss} (Section~\ref{subsec:loss}) for treating mixup generation and classification differently from a local and global perspective. At the same time, specially designed \textit{Mixer} (Section~\ref{subsec:mixblock}) generate mixed samples adaptively either at \textit{instance-level} or \textit{cluster-level} to tackle the trivial solutions effectively. 
Furthermore, in order to eliminate the drawback of poor versatility and computational overhead in SAMix optimization, we propose a pre-trained setting, SAMix$^\mathcal{P}$, which employs a pre-trained Mixer to generate high-quality mixed samples balancing performance and speed for various downstream applications. 
Noticed that SAMix$^\mathcal{P}$ can achieve competitive or slightly higher performances than SAMix in classification tasks. 
Comprehensive experiments demonstrate the effectiveness and transferring abilities of SAMix.
Our contributions are summarized as follows:
\begin{itemize}
    \vspace{-4pt}
    \item We decompose mixup learning objectives into local and global terms and further analyze the corresponding properties (local smoothness and global discrimination) of mixup generation, then design $\eta$-balanced loss to targetedly boost mixup generation performance.
    \vspace{-3pt}
    \item We build a label-free mixup generator, Mixer, with mixing attention and non-linear content modeling which effectively tackles the trivial solution problem in existing learnable methods and thus makes it more adaptive and transferable for varied scenarios.
    \vspace{-3pt}
    \item Combining the above $\eta$-balanced loss and specially tailored label-free Mixer, a unified scenario-agnostic mixup training framework, SAMix, is proposed that supports online and pre-trained pipelines for both SL and SSL tasks. 
    \vspace{-3pt}
    \item Built on SAMix framework, a pre-trained version named SAMix$^\mathcal{P}$ is provided, which brings SAMix more favorable  performance-efficiency trade-offs and generalizability across multifarious visual downstream tasks.
    %
\end{itemize}

\section{Preliminaries}
\label{sec:problem}
Given a finite set of i.i.d samples, $X=[x_i]_{i=1}^{n} \in \mathbb{R}^{D\times n}$, each data $x_{i}\in \mathbb{R}^{D}$ is drawn from a mixture of, say $C$, distributions $\mathcal{D}=\{ \mathcal{D}_{c}\}_{c=1}^C$. Our basic assumption for discriminative representations is that the each component distribution $\mathcal{D}_{c}$ has relatively low-dimensional intrinsic structures, \textit{i.e.,} the distribution $\mathcal{D}_{c}$ is constrained on a sub-manifold, say $\mathcal{M}_c$ with dimension $d_c\ll D$. The distribution $\mathcal{D}$ of $X$ is consisted of sub-manifolds, $\mathcal{M} = \cup_{c=1}^C\mathcal{M}_c$. We seek a low-dimensional representation $z_i\in \mathcal{M}$ of $x_i$ by learning a continuous mapping by a network encoder, $f_{\theta}(x):x\longmapsto z$ with the parameter $\theta\in \Theta$, which captures intrinsic structures of $\mathcal{M}$ and facilitates the discriminative tasks.

\subsection{Discriminative Representation Learning}
\label{subsec:discriminative}

\textbf{Parametric training with class supervision.}\quad
\textit{Some supervised class information} is available to ease the discriminative tasks in practical scenarios. 
Here, we assume that a one-hot label $y_i\in \mathbb{R}^C$ of each sample $x_i$ can be somehow obtained, $Y=[y_1, y_2, ..., y_n]\in \mathbb{R}^{C\times n}$. We denote the labels generated during training as pseudo labels (PL), while the fixed as ground truth labels (L). Notice that each component $\mathcal{D}_{c}$ is considered \textit{separated} according to $Y$ in this scenario. 
Then, a \textit{parametric} classifier, $g_{\omega}(z):z\longmapsto p$ with the parameter $\omega\in \Omega$, can be learned to map the representation $z_i$ of each sample to its class label $y_i$ by predicting the probability of $p_i$ being assigned to the $c$-th class using the softmax criterion, $p_{c|i} = \frac{\exp(w_{c}^T z_{i})}{\sum_{j=1}^{C}\exp(w_{j}^T z_{i})}$, where $w_c$ is a weight vector for the class $c$, and $w_{c}^{T}z_{i}$ measures how similar between $z_i$ and the class $c$. The learning objective is to minimize the cross-entropy loss (CE) between $y_{i}$ and $p_{i}$,
\begin{equation}
    \vspace{-1pt}
    \ell^{\,CE}(y_i, p_i) = - y_{i} \log p_{i}.
    \vspace{-1pt}
    \label{eq:param}
\end{equation}


\textbf{Non-parametric training as instance discrimination.}\quad
Complementary to the above parametric settings, \textit{non-parametric} approaches are usually adopted in unsupervised scenarios. Due to the \textit{lack of class information}, an instance discriminative task can be designed based on an assumption of \textit{local compactness}. 
We mainly discuss contrastive learning (CL) and take MoCo~\cite{he2020momentum} as an example. 
Consider a pair of augmented image $(x_{i}^{\tau_{q}}, x_{i}^{\tau_k})$ from the same instance $x_{i}\in \mathbb{R}^{C\times H\times W}$, the local compactness is introduced by alignment of the encoded representation pair $(z_{i}^{\tau_q},z_{i}^{\tau_k})$ from $f_{\theta,q}$ and the momentum $f_{\theta,k}$, and constrained to the global uniformity by contrasting $z_{i}^{\tau_q}$ to a dictionary of encoded keys from other images, $\{z_{j}^{\tau_k}\}_{j=1}^{K}$, where $K$ denotes the length of the dictionary. It can be achieved by the popular non-parametric CL loss with temperature $t$, called infoNCE~\cite{oord2019CPC}:
\begin{equation}
    \vspace{-3pt}
    \ell^{\,NCE}(z_{i}^{\tau_q}, z_{i}^{\tau_k}) = -\log\frac{\exp(z_{i}^{\tau_q} z_{i}^{\tau_k}/t)}{\sum^K_{j=1}\exp(z_{i}^{\tau_q} z_{j}^{\tau_k}/t)}.
    \vspace{-4pt}
    \label{eq:infonce}
\end{equation}

\subsection{Mixup for Discriminative Representation}
\label{subsec:mixup_problem} 
We further consider mixup as a generation task into discriminative representation learning to form a closed-loop framework. 
Then we have two mutually benefited sub-tasks: (a) \textit{mixed data generation} and (b) \textit{classification}. 
As for the sub-task (a), we define two functions, $h(\cdot)$ and $v(\cdot)$, to generate mixed samples and labels with a mixing ratio $\lambda \sim Beta(\alpha, \alpha)$. 
Given the mixed data, (b) defines a mixup training objective to optimize the representation space between instances or classes.

\textbf{Mixup classification as the main task.}\quad
Since we aim to seek a good representation to facilitate discriminative tasks, thus the mixed samples should be diverse and well-characterized.
The mixed samples with semantic information can be easily obtained by parametric learning, while it becomes a challenge without supervision.
Therefore, two types of the mixup classification objective $\mathcal{L}_{\theta, \omega}$ can be defined for \textit{class-level} and \textit{instance-level} mixup training. 
As for parametric training, given two randomly selected data pairs, $(x_i,y_i)$ and $(x_j,y_j)$, the mixed data is generated as $x_{m} = h(x_i, x_j, \lambda)$ and $y_{m} = v(y_i, y_j, \lambda)$. 
The objective of class-level mixup is similar to Eq.~\ref{eq:param} as, 
\begin{equation}
    \ell^{\,CE}(y_{m}, p_{m}) = \lambda \ell^{\,CE}(y_{m}, p_{m}) + (1-\lambda) \ell^{\,CE}(y_{m}, p_{m}).
    \label{eq:mixup_cls}
\end{equation}
Notice that we fix $v(\cdot)$ as the linear interpolation in our discussions, \textit{i.e.}, $v(y_i, y_j, \lambda) \triangleq \lambda y_i + (1-\lambda)y_j$. 
Symmetrically, we denote $h(\cdot)$ as a pixel-wise mixing policy with element-wise product $\odot$ for most input mixup methods~\cite{zhang2017mixup, yun2019cutmix, kim2020puzzle}, \textit{i.e.,} $x_{m} = s_{i}\odot x_{i} + s_{j}\odot x_{j}$, where $s_{i} \in \mathbb{R}^{H\times W}$ is a pixel-wise mask and $s_{j} = 1-s_{i}$. Notice that each coordinate $s_{w,h}\in [0,1]$. 
We can generate $x_{m}$ with a pair of randomly selected samples $(x_{i}, x_{j})$ and formulate mixup infoNCE loss for instance-level mixup:
\begin{equation}
    \ell^{NCE}(z_{m}) = \lambda\ell^{NCE}(z_{m}, z_{i}) + (1-\lambda)\ell^{NCE}(z_{m}, z_{j}),
    \label{eq:mixnce}
\end{equation}
where $z_{m}$, $z_{i}$ and $z_{j}$ denote the corresponding representations.
The major difference with Eq.~\ref{eq:mixup_cls} is that the augmentation view that generates $z_m$ is not from the same view, \textit{i.e.,} $z_i$ and $z_j$, as the objective function, which is effective in retaining task-relevant information, details in \ref{app_subsec:properties}.


\textbf{Mixup generation as the auxiliary task.}\quad
Unlike the learning object on the \textit{unmixed} data $X$ in Sec.~\ref{subsec:discriminative}, the performance of (b) mixup classification mainly depends on the quality of (a) mixup generation. 
We thus regard (a) as an auxiliary task to (b) and model $h(\cdot)$ as a sub-network $\mathcal{M}_{\phi}$ with the parameter $\phi\in \Phi$, called Mixer. 
Specifically, (a) aims to generate a pixel-wise mask $s\in \mathbb{R}^{H\times W}$ for mixing sample pairs. 
The mixup mask $s_{i}$ should directly related to $\lambda$ and the contents of $(x_i,x_j)$.
Practically, our $\mathcal{M}_{\phi}$ takes $l$-th layer feature maps $z^{l}\in \mathbb{R}^{C_{l}\times H_{l}\times W_{l}}$ and $\lambda$ value as the input, 
$\mathcal{M}_{\phi}: x_{i},x_{j},z_{i}^{l},z_{j}^{l},\lambda \longmapsto x_{m}$. 
The generation process of $\mathcal{M}_{\phi}$ can be trained by a mixup generation loss as $\mathcal{L}_{\phi}^{gen}$, and a mask loss designed for generated mask $s_{i}$ denoted as $\mathcal{L}_{\phi}^{mask}$. 
Formally, we have the mixup generation loss as $\mathcal{L}_{\phi} = \mathcal{L}_{\phi}^{cls} + \mathcal{L}_{\phi}^{mask}$, and the final learning objective is,
\begin{equation}
    \mathop{\min}\limits_{\theta, \omega, \phi} \mathcal{L}_{\theta, \omega} + \mathcal{L}_{\phi}.
    \label{eq:total}
\end{equation}
Both $\mathcal{L}_{\theta, \omega}$ and $\mathcal{L}_{\phi}$ can be optimized alternatively in a unified framework using a momentum pipeline with the stop-gradient operation~\cite{nips2020byol, liu2022automix}, as shown in Figure~\ref{fig:pipeline} (left).
\textbf{However, in SSL, this framework can easily make the generator fall into a trivial solution.}
To solve this problem, we propose a novel mixup loss function and a generator architecture, Mixer.
Combining the two can fully exploit the ability of mixup in learning discriminative features.

\section{SAMix for Discriminative Representations}
\label{sec:method}

\subsection{Learning Objective for Mixup Generation}
\label{subsec:loss}

Typically the objective function $\mathcal{L}_{\phi}$ corresponding to the mixup generation is consistent with the classification in parametric training (\textit{e.g.,} $\ell^{CE}$).
In this paper, we argue that mixup generation is aimed to \textit{optimize the local term subject to the global term}. 
The local term focuses on the classes of sample pairs to be mixed, while the global term introduces the constraints of other classes.
For example, $\ell^{CE}$ is the global term whose each class produces an equivalent effect on the final prediction without focusing on the relevant classes of the current sample pair.
At the class level, to emphasize the local term, we introduce a parametric binary cross-entropy (pBCE) loss for the generation task.
Formally, assuming $y_i$ and $y_j$ belong to the class $a$ and class $b$, pBCE can be summarized as:
\begin{equation}
    \ell_{+}^{\,CE}(p_{m}) = -\lambda y_{i,a}\log p_{m} - (1-\lambda) y_{j,b}\log p_{m},
    \label{ep:pBCE}
\end{equation}
where $y_{i,a}=1$ and $y_{i,b}=1$ denote the one-hot label for the class $a$ and $b$. Notice that we use $\ell_{+}$ and $\ell_{-}$ to represent the local and global terms, and $\ell_{-}$ for the parametric loss refers to $\ell^{\,CE}$. Symmetrically, we have non-parametric binary cross-entropy mixup loss (BCE) for CL:
\begin{equation}
    \ell_{+}^{NCE}(z_{m}) = -\lambda \log p_{m,i} - (1-\lambda) \log p_{m,j},
    \label{ep:BCE}
\end{equation}
where $p_{m,i}=\frac{\exp(z_{m} z_{i}/t)}{\exp(z_{m}z_i/t) + \exp(z_{m}z_j/t)}$ and its $\ell_{-}$ refers to $\ell^{NCE}$. 


\begin{figure}[b]
    \vspace{-1.0em}
    \centering
    \includegraphics[width=0.78\linewidth]{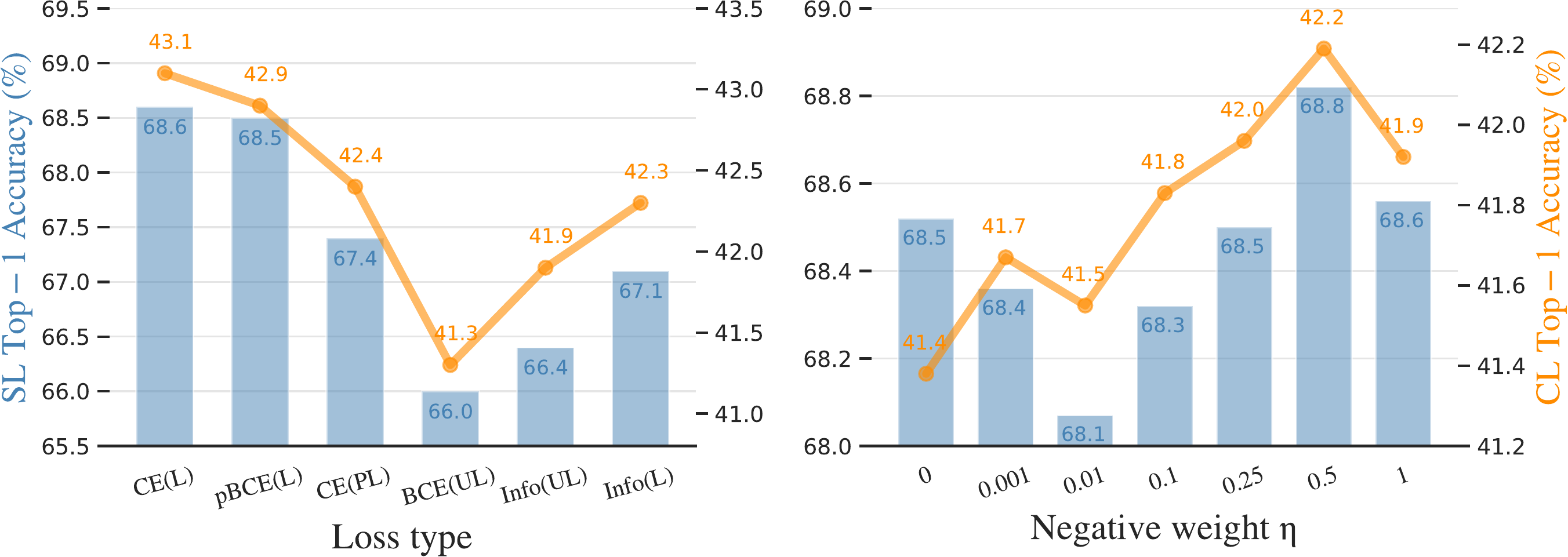}
    \vspace{-1.0em}
    \caption{Analysis of the learning objective of Mixer on Tiny ImageNet with ResNet-18. The left figure shows the results of various losses on the SL task (\textcolor{RoyalBlue}{left y axis}) and the CL task (\textcolor{YellowOrange}{right y axis}). The right one shows the effect of using various negative weights $\eta$.}
    \vspace{-1.5em}
    \label{fig:loss_analysis}
\end{figure}

\begin{figure*}[t]  
    \vspace{-3.0em}
    \centering
    \includegraphics[width=1.0\linewidth]{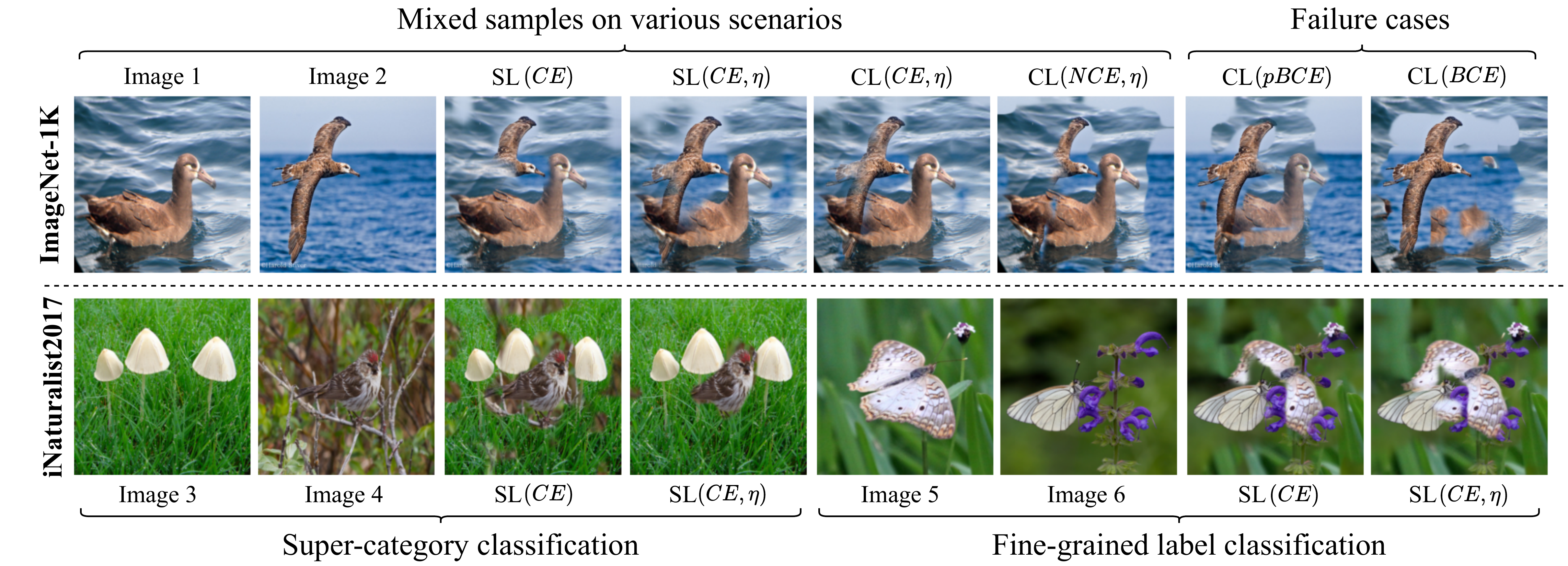}
    \vspace{-2.0em}
    \caption{Visualization and comparison of mixed samples from Mixer in various learning scenarios on IN-1k and iNat2017. Note that $\lambda=0.5$ and $\eta=0.5$ if the balance coefficient $\eta$ is included. CL(C) and CL(I) denote using SAMix-C and SAMix-I separately.}
    \label{fig:vis_main}
    \vspace{-1.5em}
\end{figure*}

\textbf{Balancing local and global terms.}\quad
Since both the local and global terms contribute to mixup generation, we analyze the importance of each term in both the SL and SSL tasks to design a balanced learning objective. 
We first analyze the properties of both terms with two hypothesizes: 
(i) the local term $\ell_{+}$ \textit{determines} the generation performance, 
(ii) the global term $\ell_{-}$ improves global discrimination but is sensitive to class information. 
To verify these properties, we design an empirical experiment based on the proposed Mixer on Tiny (see \ref{app_sec:exp_settings}). 
The main difference between the mixup CE and infoNCE is whether to adopt parametric class centroids. 
Therefore, we compare the intensity of class information among unlabeled (UL), pseudo labels (PL), and ground truth labels (L). 
Notice that PL is generated by ODC~\cite{cvpr2020odc} with the cluster number $C$. 
The class supervision can be imported to mixup infoNCE loss by filtering out negative samples with PL or L as~\cite{nips2020SupCon} denoted as infoNCE (L) and infoNCE (PL). 
As shown in Figure~\ref{fig:loss_analysis} (left), our hypothesizes are verified in the SL task (as the performance decreases from CE(L) to pBCE(L) and CE(PL) losses), but the opposite result appears in the CL task. 
The performance increases from InfoNCE(UL) to InfoNCE(L) as the false negative samples are removed~\cite{2021iclrHCL, nips2020SupCon} while trivial solutions occur using BCE(UL) (in Figure~\ref{fig:vis_main}). 
Therefore, we propose it is better to explicitly import class information as PL for instance-level mixup to generate "strong" inter-class mixed samples while preserving intra-class compactness.

\textbf{SAMix with $\eta$-balanced generation objectives.}\quad
Practically, we provide two versions of the learning objective: the mixup CE loss with PL as the class-level version (SAMix-C), and the mixup infoNCE loss as the instance-level version (SAMix-I). 
Then, we hypothesize that the best performing mixed samples will be close to the sweet spot: achieving $\lambda$ smoothness locally between two classes or neighborhood systems while globally discriminating from other classes or instances.
We propose an \textit{$\eta$-balanced mixup loss} as the objective of mixup generation,
\begin{equation}
    \ell_{\eta} = \ell_{+} + \eta \ell_{-},\ \eta \in [0,1].
    \label{eq:eta_loss}
\end{equation}
We analyze the performance of using various $\eta$ in Eq.~\ref{eq:eta_loss} on Tiny, as shown in Figure~\ref{fig:loss_analysis} (right), and find that using $\eta=0.5$ performs best on both the SL and CL tasks. In the end, we provide the learning objective, 
$\mathcal{L}_{\phi}^{cls} \triangleq \ell^{\,CE}_{+} + \eta \ell^{\,CE}_{-}$, with L for class-level mixup and with PL for SAMix-C, $\mathcal{L}_{\phi}^{cls} \triangleq \ell^{NCE}_{+} + \eta \ell^{NCE}_{-}$ for SAMix-I (details in~\ref{app_subsec:samix}).

\begin{figure}[t!]
    \vspace{-2.5em}
    \centering
    \includegraphics[width=1.0\linewidth]{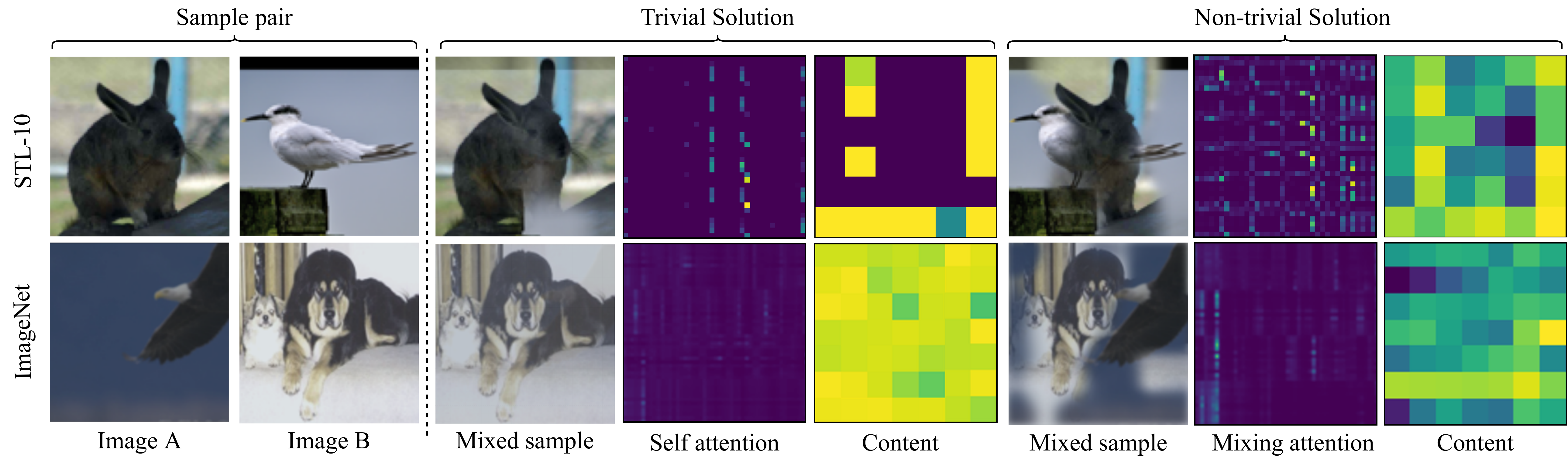}
    \vspace{-2.0em}
    \caption{Visualization of trivial (using linear $\mathcal{C}$ and self-attention) and non-trivial solutions (using the proposed non-linear $\mathcal{C}_{NC}$ and mixing attention) for SAMix-I on STL-10 and IN-1k.}
    \label{fig:trivial}
    \vspace{-1.5em}
\end{figure}

\subsection{De Novo Mixer for Mixup Generation}
\label{subsec:mixblock}
Although AutoMix proposes the MixBlock that learns adaptive mixup generation policies online, there are three drawbacks in practice: 
(a) fail to encode the mixing ratio $\lambda$ on small datasets; 
(b) fall into trivial solutions when performing CL tasks; 
(c) the online training pipeline leads to double or more computational costs than MixUp. 
Our Mixer $\mathcal{M}_{\phi}$ solves these problems individually.

\begin{wrapfigure}{r}{0.48\linewidth}
    \vspace{-3.0em}
    \begin{center}
        \includegraphics[width=1.0\linewidth]{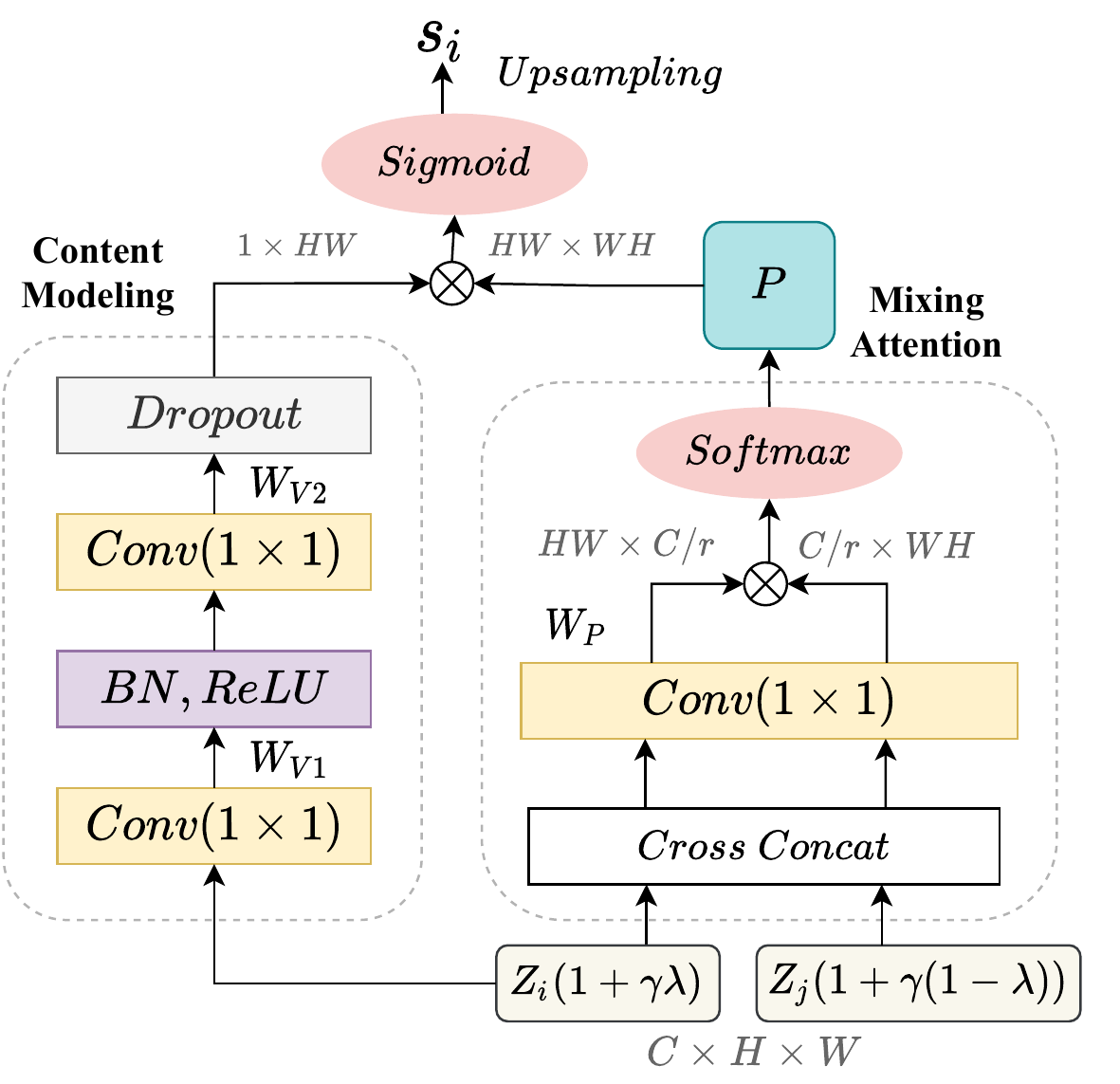}
    \end{center}
    \vspace{-1.75em}
    \caption{The network architecture of the proposed Mixer for mixup generation.}
    \label{fig:mixer}
    \vspace{-2.0em}
\end{wrapfigure}

\textbf{Adaptive $\lambda$ encoding and mixing attention.}\quad
Since a randomly sampled $\lambda$ should directly guide mixup generation, the predicted mask $s$ should be semantically proportional to $\lambda$. 
The previous typical design regards $\lambda$ as the prior knowledge and concatenates $\lambda$ to input feature maps, which might be unable to encode $\lambda$ properly (the analysis in \ref{app_subsec:mixer}). 
We propose an \textit{adaptive $\lambda$ encoding} as, 
\begin{equation}
    z^{l}_{i, \lambda} = (1+\gamma \lambda)z_i^{l},
\end{equation}
where $\gamma$ is a learnable scalar that constrained to $[0,1]$. 
Notice that $\gamma$ is initialized to $0$ during training.
As shown in Figure~\ref{fig:mixer}, we compute the mixing relationship between two samples using a new \textit{mixing attention}: we concatenate $(z^l_{i,\lambda},z^l_{j,1-\lambda})$ as the input, $\tilde z^{l} = {\rm{concat}}(z^l_{i,\lambda}, z^l_{j,1-\lambda})$, and compute the attention matrix as, 
\begin{equation}
    P_{i,j} = {\psi}(\frac{(W_{P}\tilde z^{l})^T \otimes W_{P} \tilde z^{l}}{\mathcal{N}(\tilde z^{l})}),
\end{equation}

where $\psi(\cdot)$ is the softmax function, $\mathcal{N}(\tilde z^{l})$ denotes a normalization factor, and $\otimes$ is matrix multiplication. Notice that the mixing attention provides both the cross-attention between $z^l_{i,\lambda}$ and $z^l_{j,\lambda}$ and the self-attention of each feature itself. 

\textbf{Non-linear content modeling.}\quad
In vanilla self-attention mechanism, the content sub-module $\mathcal{C}$ is a linear projection, $C_{i} = W_{z} \tilde z^{l}$, where $W_{z}$ denotes a $1\times 1$ convolution. 
However, we find the training process of this structure is unstable when performing CL tasks with the linear $\mathcal{C}$ in the early period and sometimes trapped in trivial solutions, such as all coordinates on $s_{i}$ predicted as a constant. 
As shown in Figure~\ref{fig:trivial}, we visualize $C_{i}$ and $P_{i,j}$ of trivial and non-trivial results, and find that the trivial $s_{i}$ is usually caused by a constant $C_{i}$. We hypothesize that trivial solutions happen earlier in the linear $\mathcal{C}$ than in $P_{i,j}$, because it might be unstable to project high-dimensional features to $1$-dim linearly. Hence, we design a \textit{non-linear content modeling} sub-module $\mathcal{C}_{NC}$ that contains two $1\times 1$ convolution layers with a batch normalization layer and a ReLU layer in between, as shown in Figure~\ref{fig:mixer}. 
To increase the robustness and randomness of mixup training, we add a Dropout layer with a dropout ratio of $0.1$ in $\mathcal{C}_{NC}$. Formally, Mixer $\mathcal{M}_{\phi}$ can be written as,
\begin{equation}
    s_{i} = U\bigg (
    \sigma \Big (
    {\psi}\big (\frac{(W_{P}\tilde z^{l})^T \otimes W_{P}\tilde z^{l}}{\mathcal{N}(\tilde z^{l})}\big )
    \otimes \mathcal{C}_{NC}(z^l_{i,\lambda})\Big )\bigg ).
    \label{eq:mixer}
\end{equation}

\begin{wrapfigure}{r}{0.5\linewidth}
  \vspace{-1.5em}
  \begin{center}
  \includegraphics[width=0.95\linewidth]{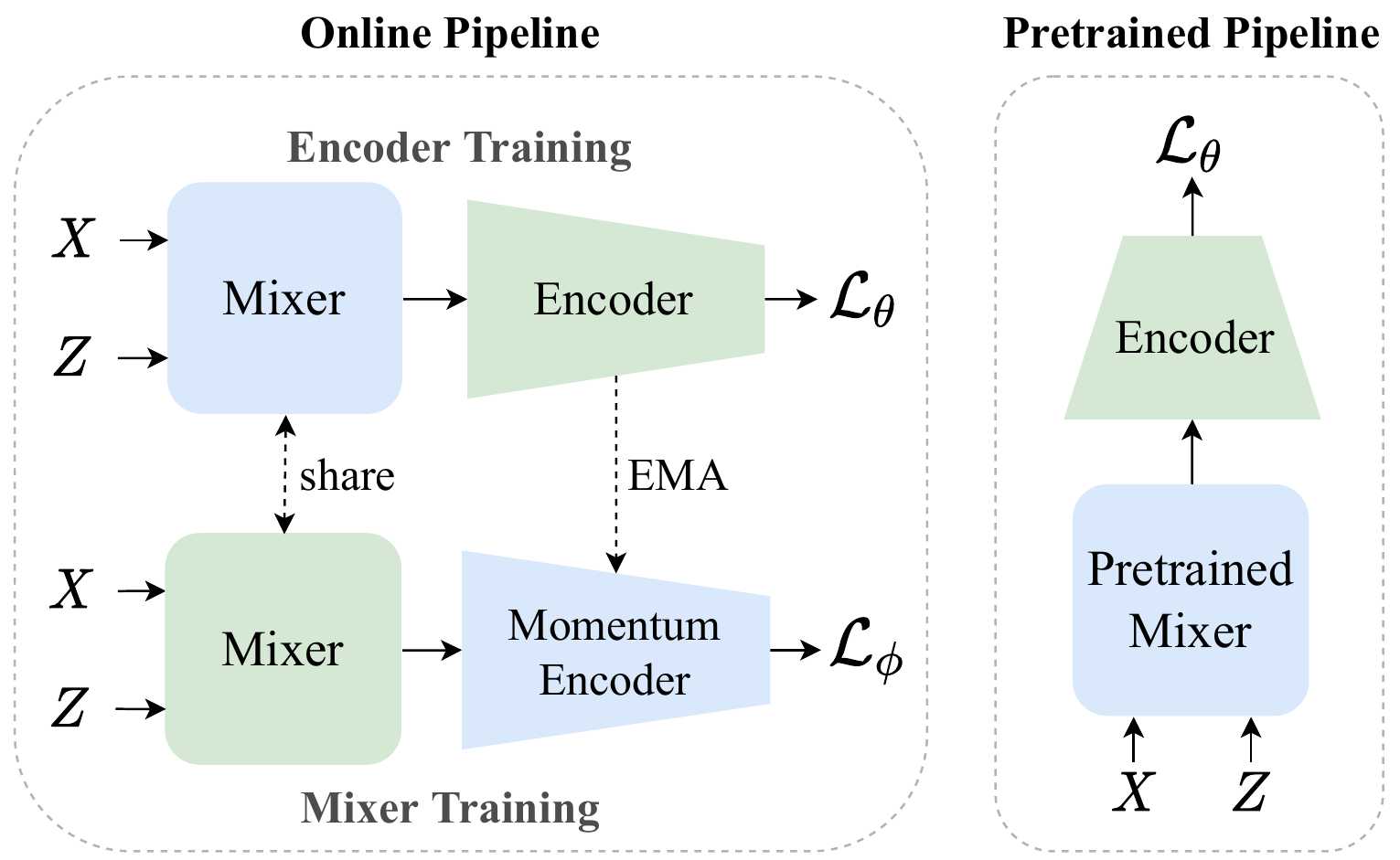}
  \end{center}
  \vspace{-1.25em}
  \caption{Training framework comparison of the online pipeline (left) in AutoMix and the proposed pre-trained pipeline (right). $X$ and $Z$ denote the input sample and corresponding feature maps from Momentum Encoder. Blue modules are not updated by back-propagation. The online pipeline optimizes Mixer and Encoder alternatively, while the pre-trained pipeline adopts pre-trained Mixer on large datasets.}
  \label{fig:pipeline}
  \vspace{-1.0em}
\end{wrapfigure}

\textbf{Pre-trained pipeline \textit{v.s.} Online pipeline.}\quad
Even though online-optimized mixup methods~\cite{liu2022automix} outperform their handcrafted counterparts by a substantial margin, their extra computational cost is intolerable especially for large datasets. 
On large-scale benchmarks, we  observe that the mixed samples generated by SAMix in both early and late training periods or with different CNN encoders vary little. Consequently, we hypothesize that, similar to knowledge distillation~\cite{dabouei2021supermix}, we can replace the online training Mixer in our SAMix with a pre-trained one. From this perspective, we propose the pre-trained SAMix pipeline, denoted as SAMix$^\mathcal{P}$, in Figure~\ref{fig:pipeline}. We can conclude: 
(i) SAMix$^\mathcal{P}$ pre-trained on large-scale datasets achieves similar or better performance as its online training version on current or relevant datasets with less computational cost. 
(ii) SAMix$^\mathcal{P}$ with light or median CNN encoders (\textit{e.g.}, ResNet-18) yields better performance than heavy ones (\textit{e.g.}, ResNet-101).
(iii) SAMix$^\mathcal{P}$ has better transferring abilities than AutoMix$^\mathcal{P}$. 
(iv) Online training pipeline is still irreplaceable on small datasets (\textit{e.g.}, CIFAR and CUB).

\textbf{Prior knowledge of mixup.}\quad
Moreover, we summarize some commonly adopted prior knowledge~\cite{kim2020puzzle, dabouei2021supermix} for mixup as two aspects: (a) adjusting the mean of $s_{i}$ correlated with $\lambda$, and (b) balancing the smoothness of local image patches while maintaining discrimination of $x_{m}$. Based on them, we introduce \textit{$\lambda$ adjusting} and modifying the mask loss $\mathcal{L}_{\phi}^{mask}$ (details in \ref{app_subsec:samix}).

\subsection{Discussion and Visualization of SAMix}
\label{subsec:discussion}
To show the influence of local and global constraints on mixup generation, we visualize mixed samples generated from Mixer on various scenarios in Figure~\ref{fig:vis_main}. It is clear that SAMix captures robust underlying data structure from both class- and instance-level effectively and thus can avoid trivial solutions and then become more applicable for SSL.

\textbf{Class-level.}\quad In the supervised task, global constraint localizes key features by discriminating to other classes, while the local term is prone to preserve more information related to the current two samples and classes. 
For example, comparing the mixed results with and without $\eta$-balanced mixup loss, it was found that pixels of the foreground target were of interest to Mixer. 
When the global constraint is balanced ($\eta=0.5$), the foreground target is retained more completely. Importantly, our designed Mixer remains invariant to the background for the more challenging fine-grained classification and preserves discriminative features.

\begin{figure*}[b!]
\vspace{-2.5em}
\centering
\begin{minipage}{0.565\linewidth}
\centering
\begin{table}[H]
    \setlength{\tabcolsep}{1.0mm}
    \caption{Top-1 Acc (\%) of image classification on IN-1k training 100-epoch and 300-epoch using procedures.}
    \vspace{-0.5em}
\resizebox{\linewidth}{!}{
    \begin{tabular}{l|ccccc|ccccc}
    \toprule
                             & \multicolumn{5}{c|}{PyTorch 100ep}                                       & \multicolumn{4}{c}{PyTorch 300ep}                         \\
    Methods                  & R-18         & R-34         & R-50         & R-101        & RX-101       & R-18         & R-34         & R-50         & R-101        \\ \hline
    Vanilla                  & 70.04        & 73.85        & 76.83        & 78.18        & 78.71        & 71.83        & 75.29        & 77.35        & 78.91        \\
    Mixup                    & 69.98        & 73.97        & 77.12        & 78.97        & 79.98        & 71.72        & 75.73        & 78.44        & 80.60        \\
    CutMix                   & 68.95        & 73.58        & 77.17        & 78.96        & 80.42        & 71.01        & 75.16        & 78.69        & 80.59        \\
    ManifoldMix              & 69.98        & 73.98        & 77.01        & 79.02        & 79.93        & 71.73        & 75.44        & 78.21        & 80.64        \\
    SaliencyMix              & 69.16        & 73.56        & 77.14        & 79.32        & 80.27        & 70.21        & 75.01        & 78.46        & 80.45        \\
    FMix$^*$                 & 69.96        & 74.08        & 77.19        & 79.09        & 80.06        & 70.30        & 75.12        & 78.51        & 80.20        \\
    PuzzleMix                & 70.12        & 74.26        & 77.54        & 79.43        & 80.63        & 71.64        & 75.84        & 78.86        & 80.67        \\
    ResizeMix$^*$            & 69.50        & 73.88        & 77.42        & 79.27        & 80.55        & 71.32        & 75.64        & 78.91        & 80.52        \\
    AutoMix                  & 70.50        & 74.52        & 77.91        & 79.87        & 80.89        & 72.05        & 76.10        & 79.25        & 80.98        \\
\rowcolor{gray90}\bf{SAMix$^\mathcal{P}$} & \blue{70.83} & \blue{74.95} & \blue{78.06} & \bf{80.05}   & \blue{80.98} & \blue{72.27} & \blue{76.28} & \blue{79.39} & \bf{81.10}   \\
\rowcolor{gray90}\bf{SAMix}               & \bf{70.85}   & \bf{74.96}   & \bf{78.11}   & \blue{80.02} & \bf{81.03}   & \bf{72.33}   & \bf{76.35}   & \bf{79.40}   & \blue{81.06} \\
    \bottomrule
    \end{tabular}
    }
    \vspace{-0.5em}
    \label{tab:sl_IN_torch}
\end{table}

\end{minipage}
\begin{minipage}{0.4275\linewidth}
\centering
\begin{table}[H]
    \setlength{\tabcolsep}{1.0mm}
    \caption{Top-1 Acc (\%) of image classification on iNat2017/2018 and Places205.}
    \vspace{-0.5em}
\resizebox{\linewidth}{!}{
    \begin{tabular}{l|cc|cc|cc}
    \toprule
                             & \multicolumn{2}{c|}{iNat2017} & \multicolumn{2}{c|}{iNat2018} & \multicolumn{2}{c}{Places205} \\
    Method                   & R-50         & RX-101         & R-50         & RX-101         & R-18         & R-50         \\ \hline
    Vanilla                  & 60.23        & 63.70          & 62.53        & 66.94          & 59.63        & 63.10        \\
    Mixup                    & 61.22        & 66.27          & 62.69        & 67.56          & 59.33        & 63.01        \\
    CutMix                   & 62.34        & 67.59          & 63.91        & 69.75          & 59.21        & 63.75        \\
    ManifoldMix              & 61.47        & 66.08          & 63.46        & 69.30          & 59.46        & 63.23        \\
    SaliencyMix              & 62.51        & 67.20          & 64.27        & 70.01          & 59.50        & 63.33        \\
    FMix$^*$                 & 61.90        & 66.64          & 63.71        & 69.46          & 59.51        & 63.63        \\
    PuzzleMix                & 62.66        & 67.72          & 64.36        & 70.12          & 59.62        & 63.91        \\
    ResizeMix$^*$            & 62.29        & 66.82          & 64.12        & 69.30          & 59.66        & 63.88        \\
    AutoMix$^*$              & 63.08        & 68.03          & 64.73        & 70.49          & 59.74        & 64.06        \\
\rowcolor{gray90}\bf{SAMix$^\mathcal{P}$} & \bf{63.38}   & \blue{68.23}   & \bf{65.16}   & \bf{70.56}     & \blue{59.82} & \bf{64.35}   \\
\rowcolor{gray90}\bf{SAMix}               & \blue{63.32} & \bf{68.26}     & \blue{64.84} & \blue{70.54}   & \bf{59.86}   & \blue{64.27} \\
    \bottomrule
    \end{tabular}
    }
    \vspace{-0.5em}
    \label{tab:sl_inat_place}
\end{table}

\end{minipage}
\vspace{-10pt}
\end{figure*}

\begin{figure*}[b!]
\vspace{-8pt}
\centering
\begin{minipage}{0.655\linewidth}
\centering
\begin{table}[H]
    \centering
    \setlength{\tabcolsep}{0.9mm}
    \caption{Top-1 Acc (\%) of supervised image classification on CIFAR-100, Tiny-ImageNet, CUB-200 and Aircraft.}
    \vspace{-0.5em}
\resizebox{\linewidth}{!}{
    \begin{tabular}{l|ccc|cc|cc|cc}
    \toprule
                   & \multicolumn{3}{c|}{CIFAR-100}             & \multicolumn{2}{c|}{Tiny-ImageNet} & \multicolumn{2}{c|}{CUB-200} & \multicolumn{2}{c}{FGVC-Aircraft}  \\
    Method         & R-18         & RX-50        & WRN-28-8     & R-18         & RX-50               & R-18         & RX-50         & R-18         & RX-50               \\ \hline
    Vanilla        & 78.04        & 81.09        & 81.63        & 61.68        & 65.04               & 77.68        & 83.01         & 80.23        & 85.10               \\
    Mixup          & 79.12        & 82.10        & 82.82        & 63.86        & 66.36               & 78.39        & 84.58         & 79.52        & 85.18               \\
    CutMix         & 78.17        & 81.67        & 84.45        & 65.53        & 66.47               & 78.40        & 85.68         & 78.84        & 84.55               \\
    ManifoldMix    & 80.35        & 82.88        & 83.24        & 64.15        & 67.30               & 79.76        & 86.38         & 80.68        & 86.60               \\
    SaliencyMix    & 79.12        & 81.53        & 84.35        & 64.60        & 66.55               & 77.95        & 83.29         & 80.02        & 84.31               \\
    FMix$^*$       & 79.69        & 81.90        & 84.21        & 63.47        & 65.08               & 77.28        & 84.06         & 79.36        & 84.85               \\
    PuzzleMix      & 80.43        & 82.57        & 85.02        & 65.81        & 66.92               & 78.63        & 84.51         & 80.76        & 86.23               \\
    ResizeMix$^*$  & 80.01        & 81.82        & 84.87        & 63.74        & 65.87               & 78.50        & 84.77         & 78.10        & 84.08               \\
    AutoMix$^*$    & \blue{82.04} & \blue{83.64} & \blue{85.16} & \blue{67.33} & \blue{70.72}        & \blue{79.87} & \blue{86.56}  & \blue{81.37} & \blue{86.69}        \\
\rowcolor{gray90}\bf{SAMix}     & \bf{82.30}   & \bf{84.42}   & \bf{85.50}   & \bf{68.89}   & \bf{72.18}          & \bf{81.11}   & \bf{86.83}    & \bf{82.15}   & \bf{86.80}          \\ 
    \bottomrule
    \end{tabular}
    }
    \vspace{-0.5em}
    \label{tab:sl_cifar_tiny_finegrained}
\end{table}

\end{minipage}
\begin{minipage}{0.335\linewidth}
\centering
\begin{table}[H]
    \setlength{\tabcolsep}{0.9mm}
    \caption{Top-1 Acc (\%) of image classification on IN-1k.}
    \vspace{-0.5em}
\resizebox{\linewidth}{!}{
    \begin{tabular}{l|cc|cc}
    \toprule
                             & \multicolumn{2}{c|}{R-50 (A3)} & DeiT-S       & Swin-T       \\
    Methods                  & MCE            & MBCE          & MCE          & MCE          \\ \hline
    Mixup+CutMix             & 76.49          & 78.08         & 79.80        & 81.20        \\
    Mixup                    & 76.01          & 77.66         & 79.65        & 81.01        \\
    CutMix                   & 76.47          & 77.62         & 79.78        & 81.23        \\
    AttentiveMix             & 76.78          & 77.46         & 80.32        & 81.29        \\
    SaliencyMix              & 76.85          & 77.93         & 79.32        & 81.37        \\
    PuzzleMix                & 77.27          & 78.02         & 79.84        & 81.47        \\
    TransMix$^{\ddagger}$    & -              & -             & 80.70        & \blue{80.80} \\
    TokenMix$^{\ddagger}$    & -              & -             & 80.80        & 80.60        \\
    AutoMix$^*$              & 77.45          & 78.34         & 80.75        & \blue{80.80} \\
\rowcolor{gray90}\bf{SAMix$^\mathcal{P}$} & \blue{77.80}   & \bf{78.73}    & \blue{80.87} & \blue{80.80} \\
\rowcolor{gray90}\bf{SAMix}               & \bf{78.33}     & \blue{78.45}  & \bf{80.94}   & \bf{81.87}   \\
    \bottomrule
    \end{tabular}
    }
    \vspace{-0.5em}
    \label{tab:sl_IN_rsb_deit}
\end{table}

\end{minipage}
\vspace{-10pt}
\end{figure*}

\textbf{Instance-level.}\quad Since no label supervision is available for SSL, the global and local terms are transformed from class to instance. Similar results are shown in the top row, and the only difference is that SAMix-C has a more precise target correspondence compared to SAMix-I via introducing class information by PL, which further indicates the importance of the information of classes. If we only focus on local relationships, Mixer can only generate mixed samples with fixed patterns (the last two results in the top row). These failure cases imply the importance of global constraints.

\vspace{-0.5em}
\section{Experiments}
\label{sec:expt}
We first evaluate SAMix for supervised learning (SL) in Sec.~\ref{exp:sl_cls} and self-supervised learning (SSL) in Sec.~\ref{exp:ssl}, and then perform ablation studies in Sec.~\ref{exp:ablation}. Nine benchmarks are used for evaluation: CIFAR-100~\cite{krizhevsky2009learning}, Tiny-ImageNet (Tiny)~\cite{2017tinyimagenet}, ImageNet-1k (IN-1k)~\cite{russakovsky2015imagenet}, STL-10~\cite{coates2011analysis}, CUB-200~\cite{wah2011caltech}, FGVC-Aircraft (Aircraft)~\cite{maji2013fine}, iNaturalist2017/2018 (iNat2017/2018)~\cite{cvpr2018inaturalist}, and Place205~\cite{nips2014place205}. All experiments are conducted with PyTorch and reported the \textit{mean of 3 trials}. SAMix uses $\alpha=2$ and the feature layer $l=3$ while SAMix$^\mathcal{P}$ is pre-trained 100 epochs with ResNet-18 (SL tasks) or ResNet-50 (SSL tasks) on IN-1k.
The \textit{median} validation top-1 Acc of the last 10 epochs is recorded.

\subsection{Evaluation on Supervised Image Classification}
\label{exp:sl_cls}
CNNs and ViTs are used as backbone networks, including ResNet (R), Wide-ResNet (WRN)~\cite{bmvc2016wrn}, ResNeXt-32x4d (RX)~\cite{xie2017aggregated}, DeiT~\cite{icml2021DeiT}, and Swin Transformer~\cite{iccv2021swin}. We use PyTorch training procedures \cite{nips2019pytorch} by default: an SGD optimizer with cosine scheduler \cite{loshchilov2016sgdr}. A special case in Table~\ref{tab:sl_IN_rsb_deit}: RSB A3 (using LAMB optimizer~\cite{iclr2020lamb} for R-50) in timm~\cite{wightman2021rsb} and DeiT (using AdamW optimizer~\cite{iclr2019AdamW} for DeiT-S and Swin-T) training recipes are fully adopted on IN-1k. MCE and MBCE denote mixup cross-entropy and mixup binary cross-entropy in RSB A3. 
For a fair comparison, grid search is performed for hyper-parameters $\alpha\in \{0.1, 0.2, 0.5, 1, 2, 4\}$ of all mixup variants. We follow hyper-parameters in original papers by default. $*$ denotes \textit{arXiv} preprint works, $\dagger$ and $\ddagger$ denote reproduced results by official codes and originally reported results, the rest are reproduced (see \ref{app_subsec:SL} and \ref{app_subsec:CL}).


\textbf{Comparison and discussion}\quad
Table~\ref{tab:sl_cifar_tiny_finegrained} shows results on small-scale and fine-grained classification tasks. SAMix consistently improves classification performances over the previous best algorithm, AutoMix, with the improved Mixer. 
Notice that SAMix significantly improved the performance of CUB-200 and Aircraft by 1.24\% and 0.78\% based on ResNet-18, and continued to expand its dominance on Tiny by bringing 1.23\% and 1.40\% improvement on ResNet-18 and ResNeXt-50. 
As for the large-scale classification task, we benchmark popular mixup methods in Table~\ref{tab:sl_IN_torch}, \ref{tab:sl_inat_place}, and \ref{tab:sl_IN_rsb_deit}, SAMix and SAMix$^\mathcal{P}$ outperform all existing methods on IN-1k, iNat2017/2018 and Places205. Especially, SAMix$^\mathcal{P}$ yields similar or better performances than SAMix with less computational cost.


\vspace{-6pt}
\subsection{Evaluation on Self-supervised Learning}
\label{exp:ssl}
Then, we evaluate SAMix on SSL tasks pre-training on STL-10, Tiny, and IN-1k. We adopt all hyper-parameter configurations from MoCo.V2 unless otherwise stated. We compared SAMix in three dimensions in CL:
(i) compare with other mixup variants, based on our proposed cross-view pipeline, and whether the predefined cluster information is given (denotes by C) or not, as shown in Table~\ref{tab:stl_ssl}. 
(ii) longitudinal comparison with CL methods that utilize mixup variants or Mixup+\textit{latent} space mixup strategies in Table~\ref{tab:imagenet_ssl}, including MoCHi~\cite{nips2020mochi}, i-Mix~\cite{2021iclrimix}, Un-Mix~\cite{2022aaaiunmix}, and WBSIM~\cite{2020bsim}, where all comparing methods are based on MoCo.V2 except SwAV~\cite{nips2020swav}.
(iii) extend SAMix$^\mathcal{P}$ and mixup variants to various CL baselines based on ResNet-50 and ViT-S in Table~\ref{tab:imagenet_ssl_extend} and Table~\ref{tab:imagenet_ssl_vit}, compared with DACL~\cite{icml2021dacl} and SDMP~\cite{cvpr2022sdmp} (using three mixup augmentations).
In these tables, $\star$ denotes our modified methods (PuzzleMix$^*$ uses PL and Inter-Intra$^{*}$ combines inter-class CutMix with intra-class Mixup, and $n$-Mix denotes the types of mixup variants used in the SSL method.

\begin{figure*}[t]
\vspace{-1.25em}
\begin{minipage}{0.505\linewidth}
\centering
    \begin{table}[H]
    \centering
    \setlength{\tabcolsep}{1.0mm}
    \caption{Top-1 Acc (\%) of linear classification pre-trained on STL-10.}
    \vspace{-0.5em}
\resizebox{\linewidth}{!}{
    \begin{tabular}{ll|cc|cc}
    \toprule
                      &                        & \multicolumn{2}{c|}{R18} & \multicolumn{2}{c}{R50} \\ 
    CL method         & Method                 & 400ep        & 800ep        & 400ep        & 800ep      \\ \hline
    MoCo.V2           & -                      & 81.50        & 85.64        & 84.89        & 89.68      \\
                      & Mixup                  & \blue{84.51} & \blue{87.93} & \blue{88.24} & \blue{92.20}  \\
                      & ManifoldMix            & 84.17        & 87.70        & 88.06        & 91.65      \\
                      & CutMix                 & 84.28        & 87.60        & 87.51        & 90.81      \\
    MoCo.V2           & SaliencyMix            & 84.33        & 87.27        & 87.35        & 90.77      \\
                      & FMix$^*$               & 84.43        & 87.68        & 88.14        & 91.56      \\
                      & ResizeMix$^*$          & 83.88        & 87.25        & 86.88        & 90.83      \\
\rowcolor{gray90}MoCo.V2 & \bf{SAMix-I}        & \bf{85.44}   & \bf{88.58}   & \bf{88.87}   & \bf{92.41}  \\ \hline
    SwAV$^{\dag}$ (C) & -                      & 81.10        & 85.56        & 84.35        & 88.79      \\
    MoCo.V2(C)        & Inter-Intra$^{\star}$  & 84.89        & 87.85        & 88.33        & \blue{92.24}  \\
    MoCo.V2(C)        & PuzzleMix$^{\star}$    & \blue{84.98} & \blue{88.07} & \blue{88.40} & 91.98      \\
\rowcolor{gray90}\bf{MoCo.V2(C)} &\bf{SAMix-C} & \bf{85.60}   & \bf{88.63}   & \bf{88.91}   & \bf{92.45} \\
    \bottomrule
    \end{tabular}
    }
    \vspace{-0.25em}
    \label{tab:stl_ssl}
\end{table}

\end{minipage}
\begin{minipage}{0.485\linewidth}
\centering
    \begin{table}[H]
    \centering
    \setlength{\tabcolsep}{0.8mm}
    \caption{Top-1 Acc (\%) of linear classification pre-trained on Tiny-ImageNet and ImageNet-1k.}
    \vspace{-0.5em}
\resizebox{\linewidth}{!}{
    \begin{tabular}{ll|cc|ccc}
    \toprule
                       &                            & \multicolumn{2}{c|}{Tiny} & \multicolumn{2}{c}{IN-1k}            \\
    CL method          & Method                     & R18             & R50             & R18           & R50          \\ \hline
    MoCo.V2            & -                          & 38.29           & 42.08           & 52.85         & 67.66        \\
    MoCo.V2            & Mixup                      & 41.24           & 46.61           & 53.03         & 68.07        \\
    MoCo.V2            & CutMix                     & 41.62           & 46.24           & 52.98         & 68.28        \\
    MoCo.V2            & SaliencyMix                & 41.14           & 46.13           & 53.06         & 68.31        \\
    MoCo.V2(C)         & PuzzleMix$^{\star}$        & 41.86           & 46.72           & 53.46         & 68.48        \\
    MoCHi$^{\dag}$     & Mixup+$latent$             & 41.78           & 46.55           & 53.12         & 68.01        \\
    i-Mix$^{\dag}$     & Mixup+$latent$             & 41.61           & 46.57           & 53.09         & 68.10        \\
    UnMix$^{\ddagger}$ & Mixup+$latent$             & -               & -               & -             & 68.60        \\
    WBSIM$^{\ddagger}$ & Mixup+CutMix               & -               & -               & -             & 68.40        \\
\rowcolor{gray90}\bf{MoCo.V2} & \bf{SAMix-I}        & 41.97           & 47.23           & \blue{53.75}  & 68.76        \\
\rowcolor{gray90}\bf{MoCo.V2} & \bf{SAMix-I$^\mathcal{P}$} & \blue{43.57}    & \bf{48.10}      & 53.72         & \blue{68.82} \\
\rowcolor{gray90}\bf{MoCo.V2(C)} & \bf{SAMix-C}            & \bf{43.68}      & \blue{47.51}    & \bf{53.93}    & \bf{68.86}   \\
    \bottomrule
    \end{tabular}
    }
    \vspace{-0.25em}
    \label{tab:imagenet_ssl}
\end{table}

\end{minipage}
\vspace{-0.75em}
\end{figure*}

\textbf{Linear Classification}\quad
Following the linear classification protocol proposed in MoCo, we train a linear classifier on top of frozen backbone features with the supervised train set. We train 100 epochs using SGD with a batch size of 256. The initialized learning rate is set to $0.1$ for Tiny and STL-10 while $30$ for IN-1k, and decay by $0.1$ at epochs 30 and 60.
As shown in Table~\ref{tab:stl_ssl}, SAMix-I outperforms all the linear mixup methods by a large margin, while SAMix-C surpasses the saliency-based PuzzleMix when PL is available. 
And SAMix-I has both global and local properties through infoNCE and BCE losses. 
Meanwhile, Table~\ref{tab:imagenet_ssl} demonstrates that both SAMix-I and SAMix-C surpass other CL methods combined with the predefined mixup. Overall, SAMix-C yields the best performance in CL tasks, indicating it provides task-relevant information with the help of PL.
Table~\ref{tab:imagenet_ssl_extend} and Table~\ref{tab:imagenet_ssl_vit} verify the generalizability of SAMix$^\mathcal{P}$ variants on popular CL baselines, which achieve comparable performances to recently proposed algorithms that combine $n$-Mix with CL.


\textbf{Downstream Tasks}\quad
\label{exp:ssl_detection}
Following the protocol in MoCo, we evaluate transferable abilities of the learned representation of comparing methods to object detection task on PASCAL VOC~\cite{2010pascalvoc} and COCO~\cite{eccv2014MSCOCO} in Detectron2~\cite{wu2019detectron2}. We fine-tune Faster R-CNN~\cite{ren2015faster} with pre-trained models on VOC~\textit{trainval07+12} and evaluate on the VOC~\textit{test2007} set. Similarly, Mask R-CNN~\cite{2017iccvmaskrcnn} is fine-tuned (2$\times$ schedule) on the COCO~\textit{train2017} and evaluated on the COCO~\textit{val2017}. SAMix still achieves comparable performance among state-of-the-art mixup methods for CL. View results in \ref{app_subsec:downstreams}.

\begin{figure*}[t]
    \vspace{-1.25em}
    \begin{minipage}{0.575\linewidth}
        \centering
        \begin{table}[H]
    \centering
    \setlength{\tabcolsep}{0.7mm}
    \caption{Top-1 Acc (\%) of linear classification of ResNet-50 pre-trained with various SSL methods on IN-1k.}
    \vspace{-0.5em}
\resizebox{\linewidth}{!}{
    \begin{tabular}{l|cccccc}
    \toprule
    Method                     & SimCLR      & MoCo.V1     & BYOL        & SwAV        & SimSiam     & MoCo.V3     \\
    PT Epoch                   & 200         & 200         & 300         & 200         & 200         & 300         \\ \hline
	-                          & 61.6        & 61.0        & 72.3        & 69.1        & 70.0        & 72.8        \\
	Mixup                      & 61.6        & 61.2        & 72.4        & 69.2        & 70.1        & 72.8        \\
	CutMix                     & 61.8        & 61.5        & 72.5        & 69.5        & 70.3        & 73.0        \\
	i-Mix$^{\dag}$(2-Mix)      & 61.7        & 61.4        & -           & 69.4        & -           & 72.8        \\
	SDMP$^{\dag}$(3-Mix)       & \blue{62.3} & \blue{61.7} & -           & -           & -           & \bf{73.5}   \\
\rowcolor{gray90}\bf{SAMix-I$^\mathcal{P}$} & 62.2        & \blue{61.7} & \blue{72.6} & \blue{69.8} & \blue{70.4} & 73.2        \\
\rowcolor{gray90}\bf{SAMix-C$^\mathcal{P}$} & \bf{62.4}   & \bf{61.9}   & \bf{72.8}   & \bf{69.9}   & \bf{70.5}   & \blue{73.4} \\
    \bottomrule
    \end{tabular}
    }
    \vspace{-0.25em}
    \label{tab:imagenet_ssl_extend}
\end{table}

    \end{minipage}
    \begin{minipage}{0.425\linewidth}
        \centering
        \begin{table}[H]
	\centering
	\setlength{\tabcolsep}{0.8mm}
	\caption{Top-1 Acc (\%) of linear classification of ViT-S pre-trained on IN-1k.}
    \vspace{-0.5em}
\resizebox{\linewidth}{!}{
    \begin{tabular}{l|ccccc}
        \toprule
        Method                     & MoCo.V2     & BYOL        & SwAV        & MoCo.V3     \\
        PT Epoch                   & 300         & 300         & 300         & 300         \\ \hline
        -                          & 72.7        & 71.4        & 73.5        & 73.2        \\
        CutMix                     & 72.6        & 71.2        & 73.6        & 73.0        \\
        i-Mix$^{\dag}$(2-Mix)      & 71.8        & -           & 73.3        & 72.7        \\
        DACL$^{\dag}$(1-Mix)       & 72.5        & -           & -           & 72.9        \\
        SDMP$^{\dag}$(3-Mix)       & \bf{72.9}   & -           & -           & \bf{73.8}   \\
\rowcolor{gray90}\bf{SAMix-I$^\mathcal{P}$} & \blue{72.8} & \blue{72.7} & \blue{73.6} & 73.4        \\
\rowcolor{gray90}\bf{SAMix-C$^\mathcal{P}$} & \bf{72.9}   & \bf{72.8}   & \bf{73.8}   & \blue{73.6} \\
        \bottomrule
    \end{tabular}
	}
	\vspace{-0.25em}
	\label{tab:imagenet_ssl_vit}
\end{table}

    \end{minipage}
    \vspace{-1.5em}
\end{figure*}

\subsection{Ablation Study}
\label{exp:ablation}
\begin{figure}[t]
    \vspace{-1.25em}
    \centering
    \includegraphics[width=1.0\linewidth]{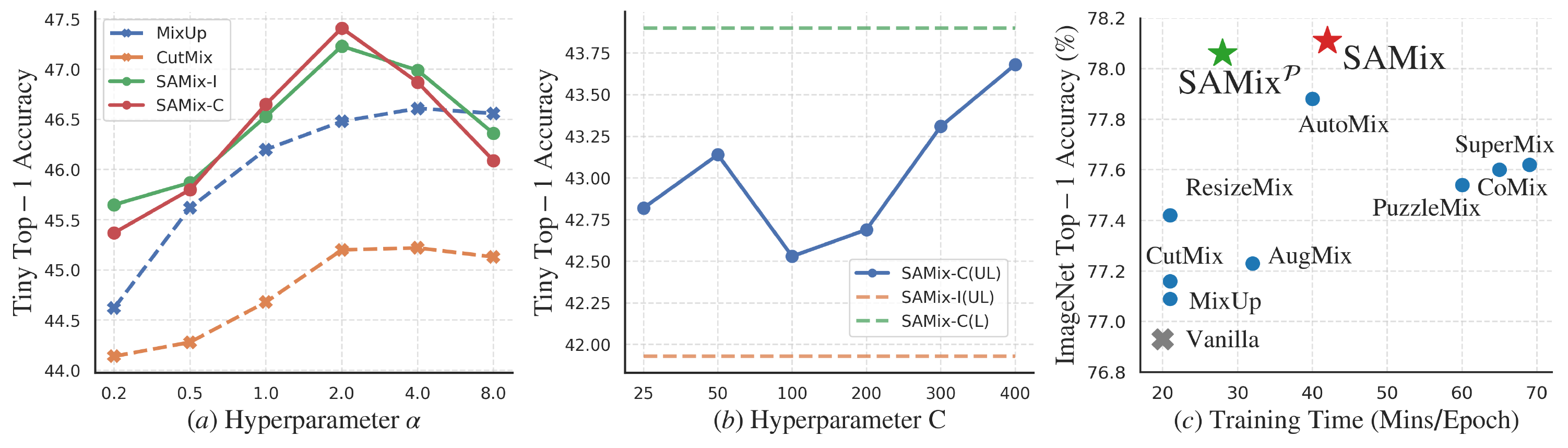}
    \vspace{-2.0em}
    \caption{(a) Hyper-parameter $\alpha$ for mixup. (b) The cluster number $C$ for SAMix-C in CL tasks on Tiny. (c) Top-1 accuracy \textit{v.s.} training time on IN-1k based on ResNet-50 with 100 epochs.}
    \label{fig:scatter}
    \vspace{-1.0em}
\end{figure}

We conduct ablation studies in four aspects: 
(i) \textbf{Mixer}: Table~\ref{tab:ablation_mixblock} verifies the effectiveness of each proposed module in both SL and CL tasks on Tiny. The first three modules enable Mixer to model the non-linear mixup relationship, while the next two modules enhance Mixer, especially in CL tasks. 
(ii) \textbf{Learning objectives}: We analyze the effectiveness of proposed $\ell_{\eta}$ with other losses, as shown in Table~\ref{tab:ablation_loss}. Using $\ell_{\eta}$ for the mixup CE and infoNCE consistently improves the performance both for the CL task on STL-10 and Tiny. 
(iii) \textbf{Time complexity analysis}: Figure~\ref{fig:scatter} (c) shows computational analysis conducted on the SL task on IN-1k using PyTorch 100-epoch settings. It shows that the overall accuracy \textit{v.s.} time efficiency of SAMix and SAMix$^\mathcal{P}$ are superior to other methods. 
(iv) \textbf{Hyper-parameters}: Figure~\ref{fig:scatter} (a) and (b) show ablation results of the hyper-parameter $\alpha$ and the clustering number $C$ for SAMix-C. We empirically choose $\alpha$=2.0 and $C=200$ as default. 

\begin{figure}[h]
\vspace{-1.5em}
\begin{center}
    \begin{minipage}{0.46\linewidth}

    \centering
        \begin{table}[H]
    \centering
    \caption{Ablation of the learning objectives in the SSL tasks on STL-10 and Tiny-ImageNet.}
\resizebox{0.90\linewidth}{!}{
    \begin{tabular}{lcc}
    \toprule
    method               & STL-10     & Tiny       \\ \hline
    BCE                  & 85.25      & 41.28      \\
    infoNCE              & 85.36      & 41.85      \\
    infoNCE ($\eta=0.5$) & \bf{85.44} & \bf{41.97} \\ \hline
    CE (PL)              & 85.56      & 42.36      \\
    CE (PL)+infoNCE      & 85.41      & 42.12      \\
    CE (PL, $\eta=0.5$)  & \bf{85.60} & \bf{42.53} \\
    \bottomrule
    \end{tabular}
    }
    \label{tab:ablation_loss}
\end{table}

    \end{minipage}
    ~
    \begin{minipage}{0.43\linewidth}
    \centering
        \begin{table}[H]
    \centering
    \caption{Ablation of proposed modules in Mixer and the $\eta$-balanced loss on Tiny.}
\resizebox{0.92\linewidth}{!}{
    \begin{tabular}{lcc}
    \toprule
    module                 & SL         & CL         \\ \hline
    Mixing attention       & 67.17      & 40.58      \\
    +Adaptive $\lambda$    & 67.95      & 41.82      \\
    +Non-linear content    & 68.46      & 42.45      \\
    +$\mathcal{L}_{mask}$  & 68.57      & 42.68      \\
    +$\lambda$ adjusting   & 68.61      & 43.14      \\ \hline
    +$\ell_{\eta}$ ($\eta=0.5$) & \bf{68.82}  & \bf{43.68} \\
    \bottomrule
    \end{tabular}
    }
    \label{tab:ablation_mixblock}
\end{table}

    \end{minipage}
\end{center}
\vspace{-2.0em}
\end{figure}

\section{Related Work}
\textbf{Class-level Mixup for SL}\quad
There are four types sample mixing policies for class-level mixup: linear mixup of input space~\cite{zhang2017mixup, yun2019cutmix, hendrycks2019augmix, harris2020fmix, qin2020resizemix} and latent space~\cite{verma2019manifold, faramarzi2020patchup}, saliency-based~\cite{uddin2020saliencymix, kim2020puzzle, kim2021comixup}, generation-based~\cite{eccv2020automix, cvpr2021alignmix}, and learning mixup generation and classification end-to-end~\cite{liu2022automix, dabouei2021supermix}. More recently, mixup designed for ViTs optimizes mixing policies with self-attention maps~\cite{2021transmix, eccv2022tokenmix}.
SAMix belongs to the fourth type and learns both class- and instance-level mixup relationships and its pre-trained SAMix$^\mathcal{P}$ eliminates high time-consuming problems of this type of method. Additionally, some researchers \cite{aaai2022grafting, 2021transmix, 2022decouplemix} improve class mixing policies upon linear mixup. See \ref{app_sec:relatedwork} for details.

\vspace{-3pt}
\textbf{Instance-level Mixup for SSL}\quad
A complementary method to learn better instance-level representation is to apply mixup in SSL scenarios \cite{2021iclrimix}. 
However, most approaches are limited to using linear mixup variants, such as applying MixUp and CutMix in the input or latent space mixup~\cite{nips2020mochi,2020bsim, icml2021dacl,cvpr2022sdmp} for SSL without ground-truth labels. 
SAMix improves SSL tasks by learning mixup policies online.

\section{Limitations and Conlusions}
In this work, we first study and decompose objectives for mixup generation as local-emphasized and global-constrained terms in order to learn adaptive mixup policy at both class- and instance-level. SAMix provides a unified mixup framework with both online and pre-trained pipelines to boost discriminative representation learning based on improved $\eta$-balanced loss and Mixer. Moreover, a more applicable pre-trained SAMix$^\mathcal{P}$ is provided. As a limitation, the Mixer only takes two samples as input and conflicts when the task-relevant information is overlapping. In future works, we suppose that k-mixup (k$\ge$2) or conflict-aware Mixer can be another promising avenue to improve mixup.

\section*{Acknowledgements}
This work was supported by National Key R\&D Program of China (No. 2022ZD0115100), National Natural Science Foundation of China Project (No. U21A20427), and Project (No. WU2022A009) from the Center of Synthetic Biology and Integrated Bioengineering of Westlake University.
This work was done when Zedong Wang and Zhiyuan Chen interned at Westlake University. We thank the AI Station of Westlake University for the support of GPUs.


\bibliographystyle{unsrt}
\bibliography{samix.bib}

\newpage
\appendix
\section{Appendix}
We first introduce dataset information in~\ref{app_subsec:basic} and provide implementation details for supervised (SL) and self-supervised learning (SSL) tasks in \ref{app_subsec:samix}, \ref{app_subsec:SL}, and \ref{app_subsec:CL}. Then, we provide settings and results of analysis experiments for Sec.~\ref{sec:method} in~\ref{app_sec:exp_settings}. Moreover, we visualize mixed samples in~\ref{app_sec:visualize}, and provide detailed related work in \ref{app_sec:relatedwork}.

\subsection{Basic Settings}
\label{app_subsec:basic}
\paragraph{Reproduction details.}
We use OpenMixup~\cite{li2022openmixup} implemented in PyTorch~\cite{nips2019pytorch} as our code-base for both supervised image classification and contrastive learning (CL) tasks. Except results marked by $\dag$ and $\ddagger$, we reproduce most experiment results of compared methods, including Mixup~\cite{zhang2017mixup}, CutMix~\cite{yun2019cutmix}, ManifoldMix~\cite{verma2019manifold}, SaliencyMix~\cite{uddin2020saliencymix}, FMix~\cite{harris2020fmix}, and ResizeMix~\cite{qin2020resizemix}.

\vspace{-12pt}
\paragraph{Dataset information.}
We briefly introduce image datasets used in Sec.~\ref{sec:expt}: 
(1) CIFAR-100~\cite{krizhevsky2009learning} contains 50k training images and 10K test images of 100 classes. (2) ImageNet-1k (IN-1k)~\cite{krizhevsky2012imagenet} contains 1.28 million training images and 50k validation images of 1000 classes. (3) Tiny-ImageNet (Tiny)~\cite{2017tinyimagenet} is a rescaled version of ImageNet-1k, which has 100k training images and 10k validation images of 200 classes. 
(4) STL-10~\cite{coates2011analysis} benchmark is designed for semi- or unsupervised learning, which consists of 5k labeled training images for 10 classes 100K unlabelled training images, and a test set of 8k images. 
(5) CUB-200-2011 (CUB)~\cite{wah2011caltech} contains over 11.8k images from 200 wild bird species for fine-grained classification. (6) FGVC-Aircraft (Aircraft)~\cite{maji2013fine} contains 10k images of 100 classes of aircraft. 
(7) iNaturalist2017 (iNat2017)~\cite{cvpr2018inaturalist} is a large-scale fine-grained classification benchmark consisting of 579.2k images for training and 96k images for validation from over 5k different wild species. 
(8) PASCAL VOC~\cite{2010pascalvoc} is a classical objection detection and segmentation dataset containing 16.5k images for 20 classes. (9) COCO~\cite{eccv2014MSCOCO} is an objection detection and segmentation benchmark containing 118k scenic images with many objects for 80 classes.

\vspace{-4pt}
\subsection{Implementation of SAMix}
\label{app_subsec:samix}
\paragraph{Online training pipeline.}
We provide the detailed implementation of SAMix in SL tasks. As shown in Figure~\ref{fig:pipeline} (left), we adopt the momentum pipeline~\cite{nips2020byol, liu2022automix} to optimize $\mathcal{L}_{\theta,\omega}$ for mixup classification and $\mathcal{L}_{\phi}$ for mixup generation in Eq.~\ref{eq:total} in an end-to-end manner:
\begin{align}
    \theta_{q}^{t},\omega_{q}^{t} &\leftarrow \mathop{{\rm argmin}}\limits_{\theta,\omega} \mathcal{L}_{\theta_{q}^{t-1}, \omega_{q}^{t-1}}, \label{eq:qem} \\
    \phi^{t} &\leftarrow \mathop{{\rm argmin}}\limits_{\phi} \mathcal{L}_{\theta_{k}^{t}, \omega_{k}^{t}} + \mathcal{L}_{\phi^{t-1}}, \label{eq:kem}
\end{align}
where $t$ is the iteration step, $\theta_{q}, \omega_{q}$ and $\theta_{k}, \omega_{k}$ denote the parameters of online and momentum networks, respectively. The parameters in the momentum networks are an exponential moving average of the online networks with a momentum decay coefficient $m$, taking $\theta_{k}$ as an example,
\begin{equation}
    \theta_{k}^{t}\leftarrow m\theta_{k}^{t-1} + (1-m)\theta_{q}^{t}.
    \label{eq:momentum}
\end{equation}
The training process of SAMix is summarized as four steps: (1) using the momentum encoder to generate the feature maps $Z^{l}$ for Mixer $\mathcal{M}_{\phi}$; (2) generating $X_{mix}^{q}$ and $X_{mix}^{k}$ by Mixer for the online networks and Mixer; (3) training the online networks by Eq.~\ref{eq:qem} and the Mixer by Eq.~\ref{eq:kem} separately; (4) updating the momentum networks by Eq.~\ref{eq:momentum}.

\vspace{-8pt}
\paragraph{Prior knowledge of mixup.}
As we discussed in Sec.~\ref{subsec:mixblock}, we introduce some prior knowledge to the Mixer from two aspects: (a) To adjust the mean of $s_{i}$ correlated with $\lambda$, we introduce a mask loss that aligns the mean of $s_{i}$ to $\lambda$, $\ell_{\mu} = \beta \max(| \lambda - \mu_{i} | - \epsilon, 0)$, where $\mu_{i} = \frac{1}{HW}\sum_{h,w} s_{i,h,w}$ is the mean and $\epsilon=0.1$ as a margin. Meanwhile, we propose a test time \textit{$\lambda$ adjusting} method. Assuming $\mu_{i}<\lambda$, we adjust each coordinate on $s_{i}$ as $\hat s_{i} = \frac{\mu_{i}}{\lambda} s_{i}$, and $\hat s_{j} = 1 - \hat s_{i}$. 
(b) To balance the smoothness of local image patches and the discrimination (e.g., variance) of $x_{m}$, we adopt a bilinear upsampling as $U(\cdot)$ for smoother masks and propose a variance loss to encourage the sparsity of learned masks, $\ell_{\sigma} = \frac{1}{WH} \sum_{w,h}(\mu_{i} - s_{w,h})^2$. We summarize the mask loss as, $\mathcal{L}_{\phi}^{mask} = \beta(\ell_{\mu} + \ell_{\sigma})$, where $\beta$ is a balancing weight. $\beta$ is initialized to $0.1$ and linearly decreases to 0 during training.

\subsection{Supervised Image Classification}
\label{app_subsec:SL}
\paragraph{Hyper-parameter settings.}
As for hyper-parameters of SAMix, we follow the basic setting in AutoMix for both SL and SSL tasks: SAMix adopts $\alpha=2$, the feature layer $l=3$, the bilinear upsampling, and the weight $\beta=0.1$ which linearly decays to $0$. We use $\eta=0.5$ for small-scale datasets (CIFAR-100, Tiny, CUB and Aircraft) and $\eta=0.1$ for large-scale datasets (IN-1k and iNat2017). 
As for other methods, PuzzleMix~\cite{kim2020puzzle}, Co-Mixup~\cite{kim2021comixup}, and AugMix~\cite{hendrycks2019augmix} are reproduced by their official implementations with $\alpha=1,2,1$ for all datasets. 
As for mixup methods reproduced by us, we provide dataset-specific hyper-parameter settings as follows. For CIFAR-100, Mixup and ResizeMix use $\alpha=1$, and CutMix, FMix and SaliencyMix use $\alpha=0.2$, and ManifoldMix uses $\alpha=2$. For Tiny, IN-1k, and iNat2017 datasets, ManifoldMix uses $\alpha=0.2$, and the rest methods adopt $\alpha=1$ for median and large backbones (e.g., ResNet-50). Specially, all these methods use $\alpha=0.2$ (only) for ResNet-18. For small-scale fine-grained datasets (CUB-200 and Aircraft), SaliencyMix and FMix use $\alpha=0.2$, and ManifoldMix uses $\alpha=0.5$, while the rest use $\alpha=1$.

\subsection{Contrastive Learning}
\label{app_subsec:CL}
\paragraph{Implementation of SAMix-C and SAMix-I.}
As for SSL tasks, we adopt the cross-view objective, $\ell^{NCE}(z_{i}^{\tau_q}, z_{i}^{\tau_k}) + \ell^{NCE}(z_{m})$, where $z_{i} = z_{i}^{\tau_k}$ and $z_j = z_{j}^{\tau_k}$, for instance-level mixup classification in all methods (except for $\dag$ and $\ddagger$ marked methods). 
We provide two variants, SAMix-C and SAMix-I, which use different learning objectives of mixup classification. The basic network structures (an encoder $f_{\theta}$ and a projector $g_{\omega}$) are adopted as MoCo.V2~\cite{2020mocov2}. Similar to SAMix in SL tasks, SAMix-C employs a parametric cluster classification head $g_{\psi}^{C}$ for online clustering~\cite{eccv2018deepcluster, cvpr2020odc} to provide pseudo labels (PL) to calculate $\mathcal{L}_{\phi}^{cls}$. It takes feature vectors from the momentum encoder as the input (optimized by Eq.~\ref{eq:kem}) and has no impact on the mixup classification objective for the online networks. Meanwhile, SAMix-I employs the instance-level classification loss for both $\mathcal{L}_{\theta,\omega}$ and $\mathcal{L}_{\phi}^{cls}$. Moreover, we use the proposed $\eta$-balanced mixup loss $\mathcal{L}_{\phi}^{cls}$ for both SAMix-C and SAMix-I with $\eta=0.5$ and the objective $\mathcal{L}_{\phi}$ for Mixer.

\paragraph{Hyper-parameter settings.}
As for Table~\ref{tab:stl_ssl} and Table~\ref{tab:imagenet_ssl}, all compared CL methods use MoCo.V2 pre-training settings except for SwAV~\cite{nips2020swav}, which adopts ResNet-50~\cite{cvpr2016resnet} as the encoder $f_{\theta}$ with two-layer MLP projector $g_{\omega}$ and is optimized by SGD optimizer and Cosine scheduler with the initial learning rate of $0.03$ and the batch size of $256$. The length of the momentum dictionary is 65536 for IN-1k and 16384 for STL-10 and Tiny datasets. The data augmentation strategy is based on IN-1k in MoCo.v2 as follows: Geometric augmentation is \texttt{RandomResizedCrop} with the scale in $[0.2,1.0]$ and \texttt{RandomHorizontalFlip}. Color augmentation is \texttt{ColorJitter} with \{brightness, contrast, saturation, hue\} strength of $\{0.4, 0.4, 0.4, 0.1\}$ with a probability of $0.8$, and \texttt{RandomGrayscale} with a probability of $0.2$. Blurring augmentation uses a square Gaussian kernel of size $23\times 23$ with a std uniformly sampled in $[0.1, 2.0]$. We use 224$\times$224 resolutions for IN-1k and 96$\times$96 resolutions for STL-10 and Tiny datasets.
As for Table~\ref{tab:imagenet_ssl_extend} and Table~\ref{tab:imagenet_ssl_vit}, we follow the original setups of these CL baselines (SimCLR~\cite{icml2020simclr}, MoCo.V1~\cite{cvpr2020moco}, MoCo.V2~\cite{2020mocov2}, BYOL~\cite{nips2020byol}, SwAV~\cite{nips2020swav}, SimSiam~\cite{cvpr2021simsiam}, and MoCo.V3~\cite{iccv2021mocov3}) using OpenMixup~\cite{li2022openmixup} implementations. Notice that MoCo.V3 is specially designed for vision Transformers \cite{iclr2021vit} while other CL baselines are originally proposed with CNN architecture. Meanwhile, we employ the contrastive learning objectives with mixing augmentations for BYOL and SimSiam proposed in BSIM~\cite{2020bsim} because these CL baselines adopt the MSE loss between positive sample pairs instead of the infoNCE loss (Eq.~\ref{eq:infonce}).

\paragraph{CL methods with Mixup augmentations.}
In Sec.~\ref{exp:ssl}, we compare the proposed SAMix variants with general Mixup approaches proposed in SL and well-designed CL methods with Mixups. As for the general Mixup variants implemented with CL baselines, Mixup~\cite{zhang2017mixup}, ManifoldMix~\cite{verma2019manifold}, CutMix~\cite{yun2019cutmix}, SaliencyMix~\cite{uddin2020saliencymix}, FMix~\cite{harris2020fmix}, ResizeMix~\cite{qin2020resizemix}, PuzzleMix~\cite{kim2020puzzle}, and out proposed SAMix only use the single Mixup augmentation. As for the CL methods applying Mixup augmentations, DACL~\cite{icml2021dacl} employs the vanilla Mixup, MoCHi~\cite{nips2020mochi}, i-Mix~\cite{2021iclrimix}, UnMix~\cite{2022aaaiunmix}, WBSIM~\cite{2020bsim} use two types of Mixup strategies in the input image and the latent space of the encoder, SDMP~\cite{cvpr2022sdmp} randomly applies three types of input space Mixups (Mixup, CutMix, and ResizeMix). Therefore, SDMP can achieve competitive performances as SAMix variants in Table~\ref{tab:imagenet_ssl_extend} and Table~\ref{tab:imagenet_ssl_vit}. 

\paragraph{Evaluation protocols.}
We evaluate the SSL representation with a linear classification protocol proposed in MoCo~\cite{he2020momentum} and MoCo.V3~\cite{iccv2021mocov3} for ResNet and ViT variants, which trains a linear classifier on top of the frozen representation on the training set. For ResNet variants, the linear classifier is trained 100 epochs by an SGD optimizer with the SGD momentum of $0.9$ and the weight decay of $0$. We set the initial learning rate of $30$ for IN-1k as MoCo, and $0.1$ for STL-10 and Tiny with a batch size of 256. The learning rate decays by $0.1$ at epochs 60 and 80. For ViT-S, the linear classifier is trained 90 epochs by the SGD optimizer with a batch size of 1024 and a basic learning rate of 12.
Moreover, we adopt object detection task to evaluate transfer learning abilities following MoCo, which uses the $4$-th layer feature maps of ResNet (ResNet-C4) to fine-tune Faster R-CNN~\cite{ren2015faster} with 24k iterations on the \textit{trainval07+12} set and Mask R-CNN~\cite{2017iccvmaskrcnn} with 2$\times$ training schedule (24-epoch) on the \textit{train2017} set.

\begin{figure}[t]
    \vspace{-1.5em}
    \centering
    \includegraphics[width=1.0\linewidth]{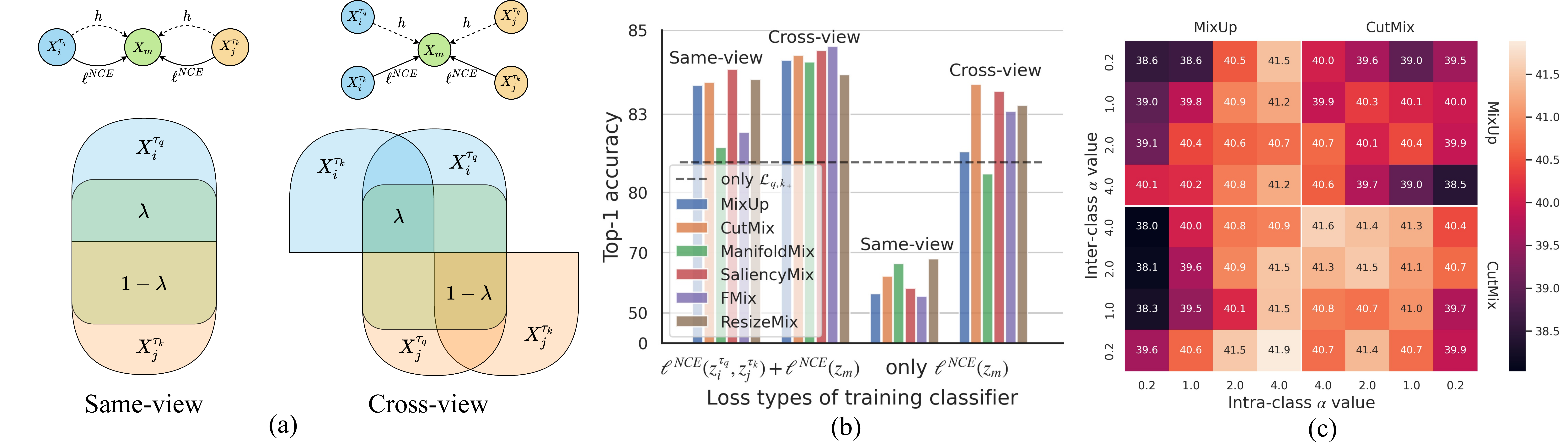}
    \vspace{-2.5em}
    \caption{
    (a) Graphical models and information diagrams of \textit{same-view} and \textit{cross-view} training pipeline for instance-level mixup. Taking the \textit{cross-view} as an example, $x_{m} = h(x_{i}^{\tau_2}, x_{j}^{\tau_2}, \lambda)$, the $\lambda$ region denotes the corresponding information partition for $\lambda I(z_{m}^{\tau_q}, z_{i}^{\tau_k})$ and the $1-\lambda$ region for $(1-\lambda) I(z_{m}^{\tau_q}, z_{j}^{\tau_k})$. (b) Linear evaluation (top-1 accuracy on STL-10) of whether to use the cross-view pipeline and to combine the original infoNCE loss with the mixup infoNCE loss. (c) A heat map of linear evaluation (top-1 accuracy on Tiny) represents the effects of using MixUp and CutMix as the inter-class (y-axis) and intra-class mixup (x-axis) using various $\alpha$.
    }
    \vspace{-10pt}
    \label{fig:app_mixup_cls}
\end{figure}

\subsection{Empirical Experiments}
\label{app_sec:exp_settings}
\subsubsection{Cross-view Training Pipeline for Instance-level Mixup}
\label{app_subsec:properties}
We first analyze the learning objective of instance-level mixup classification for contrastive learning (CL) as discussed in Sec.~\ref{subsec:mixup_problem}.
As shown in Figure~\ref{fig:app_mixup_cls} (a), there are two possible objectives for instance-level mixup defined in Eq.~\ref{eq:mixnce}: the \textit{same-view}, $\mathop{\max}\limits_{\theta, \omega} \lambda I(z_{m}^{\tau_q}, z_{i}^{\tau_q}) + (1-\lambda) I(z_{m}^{\tau_q}, z_{j}^{\tau_q})$, and \textit{cross-view} objective, $\mathop{\max}\limits_{\theta, \omega} \lambda I(z_{m}^{\tau_q}, z_{i}^{\tau_k}) + (1-\lambda) I(z_{m}^{\tau_q}, z_{j}^{\tau_k})$.
We hypothesize that the cross-view objective yields better CL performance than the same-view because the mutual information between two augmented views should be reduced while keeping task-relevant information~\cite{nips2020infomin, iclr2021ssl_multiview}. To verify this hypothesis, we design an experiment of various mixup methods with $\alpha=1$ on STL-10 with ResNet-18. As shown in Figure~\ref{fig:app_mixup_cls} (b), we compare using the same-view or cross-view pipelines combined with using $\ell^{NCE}(z_{i}^{\tau_q}, z_{i}^{\tau_k}) + \ell^{NCE}(z_{m})$ or only using $\ell^{NCE}(z_{m})$. We can conclude: 
(i) Degenerated solutions occur when using the same-view pipeline while using the cross-view pipeline outperforms the CL baseline. It is mainly caused by degenerated mixed samples which contain parts of the same view of two source images. Therefore, we propose the cross-view pipeline for the instance-level mixup, where $z_{i}$ and $z_{j}$ in Eq.~\ref{eq:mixnce} are representations of $x_{i}^{\tau_{k}}$ and $x_{j}^{\tau_{k}}$. 
(ii) Combining both the original and mixup infoNCE loss, $\ell^{NCE}(z_{i}^{\tau_q}, z_{i}^{\tau_k}) + \ell^{NCE}(z_{m})$, surpasses only using one of them, which indicates that mixup enables $f_{\theta}$ to learn the relationship between local neighborhood systems.

\subsubsection{Analysis of Instance-level Mixup}
\label{app_subsec:instance_mix}
As we discussed in Sec.~\ref{app_subsec:properties}, we propose the cross-view training pipeline for instance-level mixup classification. We then discuss inter- and intra-class proprieties of instance-level mixup. As shown in Figure~\ref{fig:app_mixup_cls} (c), we adopt inter-cluster and intra-cluster mixup from \{Mixup, CutMix\} with $\alpha \in \{0.2, 1, 2, 4\}$ to verify that instance-level mixup should treat inter- and intra-class mixup differently. Empirically, mixed samples provided by Mixup preserve global information of both source samples (smoother), while samples generated by CutMix preserve local patches (more discriminative). And we introduce pseudo labels (PL) to indicate different clusters by clustering method ODC~\cite{cvpr2020odc} with the class (cluster) number $C$. Based on experiment results, we can conclude that inter-class mixup requires \textit{discriminative} mixed samples with \textit{strong} intensity while the intra-class needs \textit{smooth} samples with \textit{low} intensity. 
Moreover, we provide two cluster-based instance-level mixup methods in Table~\ref{tab:stl_ssl} and \ref{tab:imagenet_ssl} (denoting by $*$): (a) Inter-Intra$^*$. We use CutMix with $\alpha \ge 2$ as inter-cluster mixup and Mixup with $\alpha=0.2$ as an intra-cluster mixup. (b) PuzzleMix$^*$. We introduce saliency-based mixup methods to SSL tasks by introducing PL and a parametric cluster classifier $g_{\psi}^{C}$ after the encoder. This classifier $g_{\psi}^{C}$ and encoder $f_{\theta}$ are optimized online like AutoMix and SAMix mentioned in~\ref{app_subsec:samix}. Based on Grad-CAM~\cite{selvaraju2017gradcam} calculated from the classifier, PuzzleMix can be adopted on SSL tasks.

\subsubsection{Analysis of Mixup Generation Objectives}
\label{app_subsec:gen_loss}
In Sec.~\ref{subsec:loss}, we design experiments to analyze various losses for mixup generation in Figure~\ref{fig:loss_analysis} (left) and the proposed $\eta$-balanced loss in Figure~\ref{fig:loss_analysis} (right) for both SL and SSL tasks with ResNet-18 on STL-10 and Tiny. 
Basically, we assume both STL-10 and Tiny datasets have 200 classes on their 100k images. Since STL-10 does not provide ground truth labels (L) for 100k unlabeled data, we introduce PL generated by a supervised pertained classifier on Tiny as the "ground truth" for its 100k training set. Notice that L denotes ground truth labels and PL denotes pseudo labels generated by ODC~\cite{cvpr2020odc} with $C=200$.

As for the SL task, we use the labeled training set for mixup classification (100k on Tiny \textit{v.s.} 5k on STL-10). Notice that SL results are worse than using SSL settings on STL-10, since the SL task only trains a randomly initialized classifier on 5k labeled data. Because the infoNCE and BCE loss require cross-view augmentation (or they will produce trivial solutions), we adopt MoCo.V2 augmentation settings for these two losses when performing the SL task. Compared to CE (L), we corrupt the global term in CE as CE (PL) or directly remove them as pBCE (L) to show that pBCE is vital to optimizing mixed samples. Similarly, we show that the global term is used as the global constraint by comparing BCE (UL) with infoNCE (UL), infoNCE (PL), and infoNCE (L).

As for the SSL task, we verify the conclusions drawn from the SL task and conclude that (a) the local term optimizes mixup generation directly, corresponding to the smoothness property, and (b) the global term serves as the global constraint corresponding to the discriminative property. Moreover, we verified that using the $\eta$-balanced loss as $\mathcal{L}_{\phi}^{cls}$ yields the best performance on SL and SSL tasks. Notice that we use $\eta=0.5$ on small-scale datasets and $\eta=0.1$ on large-scale datasets for SL tasks, and use $\eta=0.5$ for all SSL tasks.

\begin{table}[ht]
    \centering
    \caption{Transferring to object detection with Faster R-CNN on VOC and Mask R-CNN on COCO.}
\resizebox{0.80\linewidth}{!}{
\begin{tabular}{ll|ccc|ccc}
    \toprule
                       &                            & \multicolumn{3}{c|}{Faster R-CNN}       & \multicolumn{3}{c}{Mask R-CNN}              \\
    CL Method          & Methods                    & AP          & AP$_{50}$   & AP$_{75}$   & AP$^{b}$    & AP$_{50}^{b}$ & AP$_{75}^{b}$ \\ \hline
    MoCo.V2            & -                          & 56.9        & 82.2        & 63.4        & 40.6        & 60.1          & 44.0          \\
    MoCo.V2            & Mixup                      & 57.4        & 82.5        & 64.0        & 41.0        & 60.8          & 44.3          \\
    MoCo.V2            & CutMix                     & 57.3        & 82.7        & 64.1        & 41.1        & 60.8          & 44.4          \\
    MoCo.V2            & Inter-Intra$^{\star}$      & 57.5        & 82.8        & 64.2        & 41.2        & 60.9          & 44.4          \\
    MoCHi$^{\dag}$     & $input$+$latent$           & 57.1        & 82.7        & 64.1        & 41.0        & 60.8          & 44.5          \\
    i-Mix$^{\dag}$     & $input$+$latent$           & 57.5        & 82.7        & 64.2        & -           & -             & -             \\
    UnMix$^{\ddagger}$ & $input$+$latent$           & \blue{57.7} & 83.0        & \blue{64.3} & \blue{41.2} & 60.9          & \bf{44.7}     \\
    WBSIM$^{\ddagger}$ & $input$                    & 57.4        & 82.8        & 64.2        & 40.7        & 60.8          & 44.2          \\
    \bf{MoCo.V2}       & \bf{SAMix-I}               & 57.5        & \blue{83.1} & 64.2        & \blue{41.2} & \blue{61.0}   & 44.5          \\
    \bf{MoCo.V2}       & \bf{SAMix-I$^\mathcal{P}$} & \bf{57.8}   & \bf{83.2}   & \blue{64.3} & \bf{41.3}   & \bf{61.1}     & \blue{44.6}   \\
    \bf{MoCo.V2}       & \bf{SAMix-C}               & \blue{57.7} & \blue{83.1} & \bf{64.4}   & \bf{41.3}   & \bf{61.1}     & \bf{44.7}     \\
    \bottomrule
    \end{tabular}
    }
    \label{tab:detection}
    \vspace{-0.5em}
\end{table}

\subsubsection{Analysis of Mutual Information for Mixup}
\label{app_subsec:MI}
Since mutual information (MI) is usually adopted to analyze contrastive-based augmentations~\cite{eccv2020CMC, nips2020infomin}, we estimate MI between $x_{m}$ of various methods and $x_{i}$ by MINE~\cite{icml2018mine} with 100k images in 64$\times$64 resolutions on Tiny-ImageNet. We sample $\lambda=$ from 0 to 1 with the step of $0.125$ and plot results in Figure~\ref{fig:scatter} (d). Here we see that SAMix-C and SAMix-I with more MI when $\lambda \approx 0.5$ perform better.

\subsubsection{Results of Downstream tasks}
\label{app_subsec:downstreams}
In Sec.~\ref{exp:ssl_detection}, we evaluate transferable abilities of the learned representation of self-supervised methods to object detection task on PASCAL VOC~\cite{2010pascalvoc} and COCO~\cite{eccv2014MSCOCO}. In Table~\ref{tab:detection}, the online SAMix-C and the pre-trained SAMix-I$^\mathcal{P}$ achieve the best detection performances among the compared methods and significantly improves the baseline MoCo.V2 (\textit{e.g.,} SAMix-C gains 0.9\% AP and +0.7\% AP$^{b}$ over MoCo.V2). Notice that MoCHi, i-Mix, and UnMix introduce mixup augmentations in both the input and latent spaces, while our proposed SAMix only generates mixed samples in the input space.

\subsection{Visualization of SAMix}
\label{app_sec:visualize}
\subsubsection{Mixing Attention and Content in Mixer}
\label{app_subsec:mixer}
In Sec.~\ref{subsec:mixblock}, we discuss the trivial solutions of Mixer, which are usually occurred in SSL tasks. Given the sample pair $(x_i,x_j)$ and $\lambda=0.5$, we visualize the content $C_{i}$ and $P_{i,j}$ to compare the trivial and non-trivial results in the SSL task on STL-10, as shown in Figure~\ref{fig:trivial}. As we can see, both $C_{i}$ and $P_{i,j}$ from the trivial solution has extremely large or small scale values while $C_{i}$ generated by $C_{NCL}$ containing more balanced values. Since the attention weight $P_{i,j}$ is normalized by softmax, we hypothesize that $ C_i$ more likely causes trivial solutions. To verify our hypothesis, we freeze $W_{P}$ in the original MB and compare the original linear content projection $W_{z}$ with the non-linear content modeling. The results confirm that the non-linear module can prevent large-scale values on $C_{i}$ and eliminate the trivial solutions.

\begin{figure}[t]
    \centering
    \includegraphics[width=0.98\linewidth]{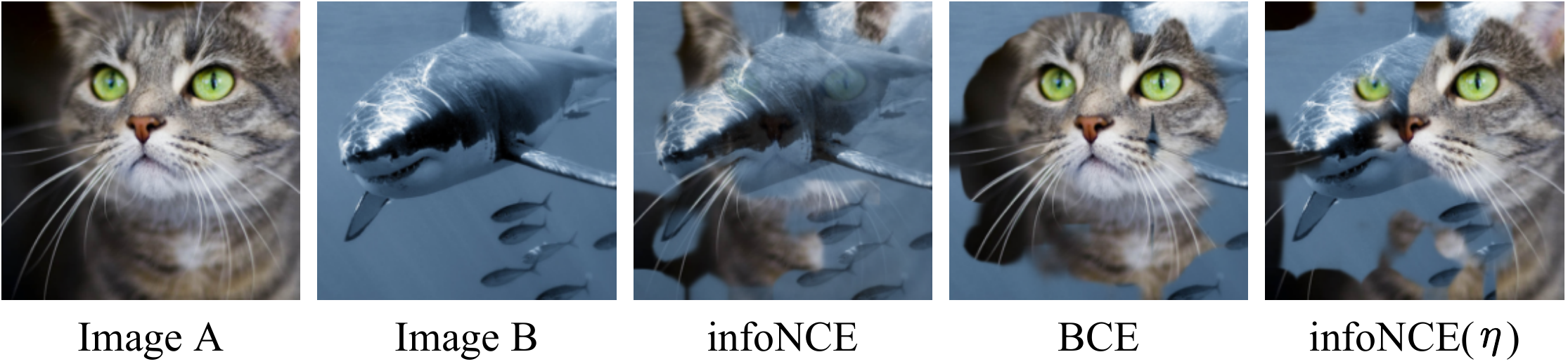}
    \vspace{-8pt}
    \caption{Visualization of loss effect. Both infoNCE and BCE loss have different emphases: infoNCE shows a similar effect of supervised fine-grained classification, focusing on fragmented and essential features, while BCE focuses on object completeness.}
    \label{fig:app_loss}
    \vspace{-10pt}
\end{figure}

\vspace{-8pt}
\subsubsection{Effects of Mixup Generation Loss}
In addition to Sec.\ref{subsec:discussion}, we further provide visualization of mixed samples using the infoNCE (Eq.~\ref{eq:mixnce}), BCE (Eq.~\ref{ep:BCE}), and $\eta$-balanced infoNCE loss (Eq.~\ref{eq:eta_loss}) for Mixer. As shown in Figure~\ref{fig:app_loss}, we find that mixed samples using infoNCE mixup loss prefer instance-specific and fine-grained features. On the contrary, mixed samples of the BCE loss seem only to consider discrimination between two corresponding neighborhood systems. It is more inclined to maintain the continuity of the whole object relative to infoNCE. Thus, combining both the characteristics, the $\eta$-balanced infoNCE loss yields mixed samples that retain both instance-specific features and global discrimination.

\vspace{-6pt}
\subsubsection{Visualization of Mixed Samples in SAMix}
\label{app_subsec:mix_sample}
\begin{figure}[b]
    \vspace{-1em}
    \centering
    \includegraphics[width=0.98\linewidth]{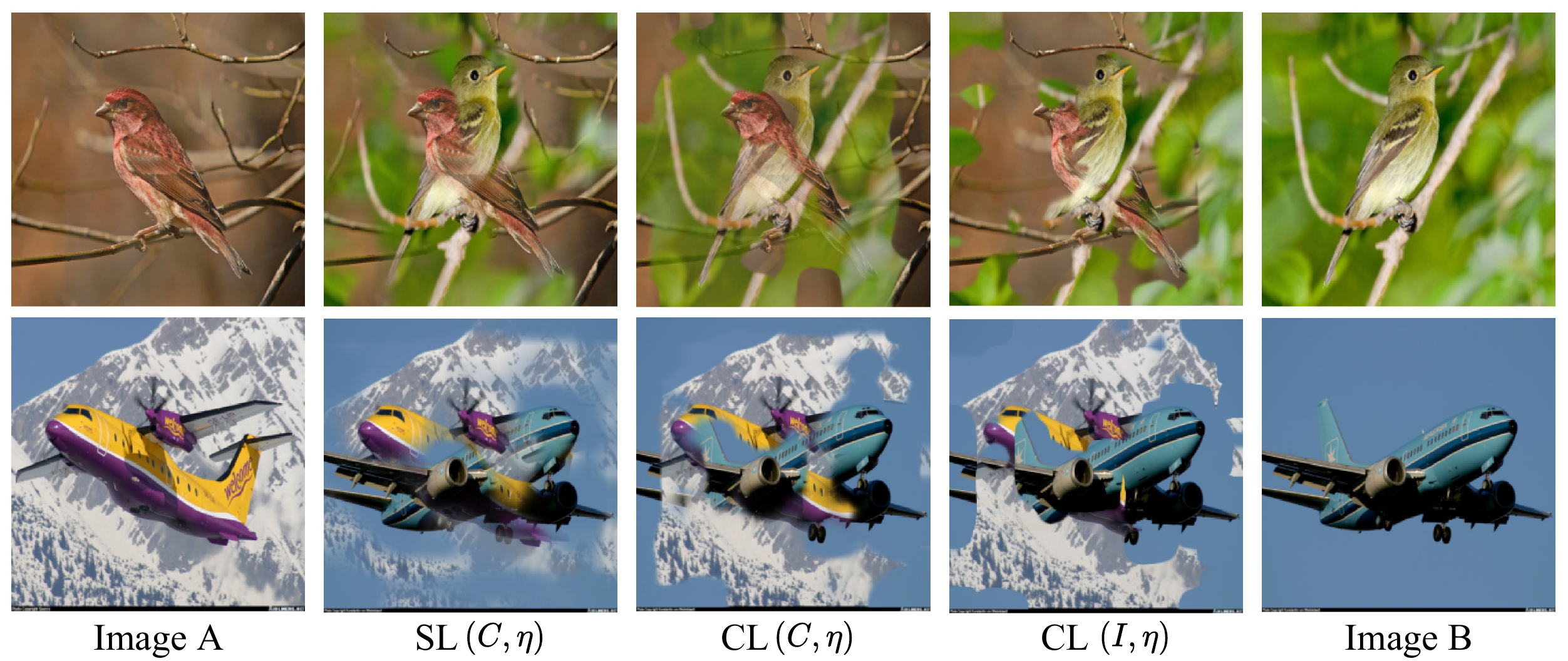}
    \vspace{-10pt}
    \caption{Visualization of SAMix in various scenarios on CUB and Aircraft. Given images A and B, the middle three mixed samples are generated by SAMix with $\lambda=0.5$ trained in the fine-grained SL task and the SSL tasks (SAMix-C and SAMix-I).}
    \label{fig:app_scenarios}
\end{figure}

\paragraph{SAMix in various scenarios.}
In addition to Sec.~\ref{subsec:discussion}, we visualize the mixed samples of SAMix in various scenarios to show the relationship between mixed samples and class (cluster) information. Since IN-1k contains some samples in CUB and Aircraft, we choose the overlapped samples to visualize SAMix trained for the fine-grained SL task (CUB and Aircraft) and SSL tasks (SAMix-I and SAMix-C). As shown in Figure~\ref{fig:app_scenarios}, mixed samples reflect the granularity of class information adopted in mixup training. Specifically, we find that mixed samples using infoNCE mixup loss (Eq.\ref{eq:mixnce}) is more closely to the fine-grained SL because they both have many fine-grained centroids.

\begin{figure*}[t]
    \centering
    \includegraphics[width=0.98\linewidth]{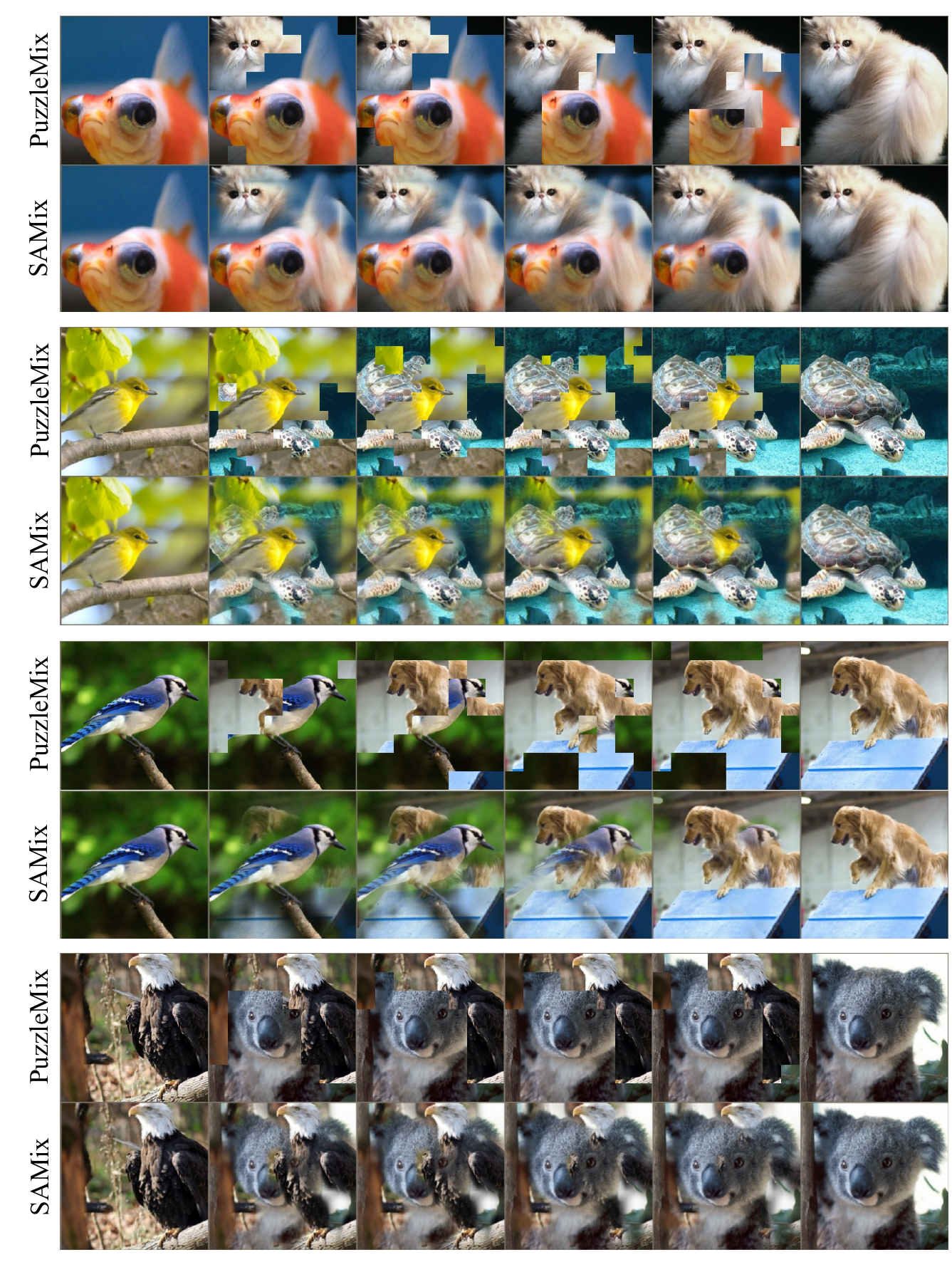}
    \vspace{-1.0em}
    \caption{Visualization of PuzzleMix \textit{v.s.} SAMix for SL tasks on IN-1k. In every four rows, the upper and lower two rows represent mixed samples generated by PuzzleMix and SAMix, respectively. $\lambda$ value changes from left ($\lambda=0$) to right ($\lambda=1$) by an equal step.}
    \label{fig:SL_AutoMix_SAMix}
\end{figure*}

\vspace{-8pt}
\paragraph{Comparison with PuzzleMix in SL tasks.}
To highlight the accurate mixup relationship modeling in SAMix compared to PuzzleMix (standing for saliency-based methods), we visualize the results of mixed samples from these two methods in the supervised case in Figure~\ref{fig:SL_AutoMix_SAMix}. There is three main difference: (a) bilinear upsampling strategy in SAMix makes the mixed samples more smooth in local patches. (b) adaptive $\lambda$ encoding and mixing attention enhances the correspondence between mixed samples and $\lambda$ value. (c) $\eta$-balanced mixup loss enables SAMix to balance global discriminative and fine-grained features.

\begin{figure*}[t]
    \centering
    \includegraphics[width=0.98\linewidth]{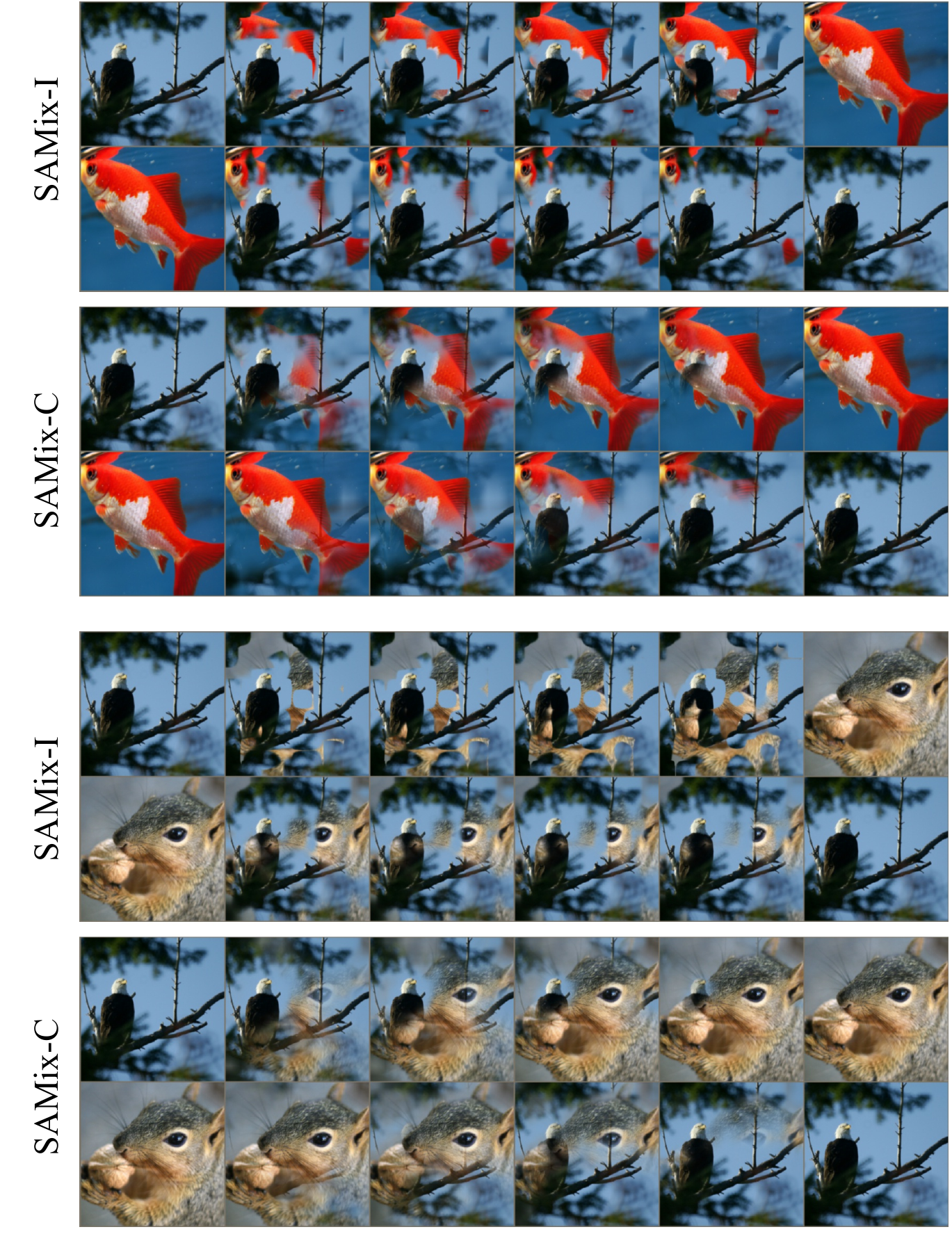}
    \vspace{-0.5em}
    \caption{Visualization of SAMix-I \textit{v.s.} SAMix-C for SSL tasks on IN-1k. In every four rows, the upper and lower two rows represent mixed samples generated by SAMix-I and SAMix-C, respectively. $\lambda$ value changes from left ($\lambda=0$) to right ($\lambda=1$) by an equal step.}
    \label{fig:SSL_SAMIX_I_C}
\end{figure*}

\vspace{-8pt}
\paragraph{Comparison of SAMix-I and SAMix-C in SSL tasks.}
As shown in Figure~\ref{fig:SSL_SAMIX_I_C}, we provide more mixed samples of SAMix-I and SAMix-C in the SSL tasks to show that introducing class information by PL can help Mixer generate mixed samples that retain both the fine-grained features (instance discrimination) and whole targets.


\subsection{Detailed related work}
\label{app_sec:relatedwork}
\paragraph{Contrastive Learning.} CL amplifies the potential of SSL by achieving significant improvements on classification~\cite{icml2020simclr, he2020momentum, nips2020swav}, which maximizes similarities of positive pairs while minimizing similarities of negative pairs. To provide a global view of CL, MoCo~\cite{he2020momentum} proposes a memory-based framework with a large number of negative samples and model differentiation using the exponential moving average. SimCLR~\cite{icml2020simclr} demonstrates a simple memory-free approach with large batch size and strong data augmentations that is also competitive in performance to memory-based methods. BYOL~\cite{nips2020byol} and its variants~\cite{cvpr2021simsiam, iccv2021mocov3} do not require negative pairs or a large batch size for the proposed pretext task, which tries to estimate latent representations from the same instance. Besides pairwise contrasting, SwAV~\cite{nips2020swav} performs online clustering while enforcing consistency between multi-views of the same image. Barlow Twins~\cite{icml2021barlow} avoids the representation collapsing by learning the cross-correlation matrix of distorted views of the same sample. Moreover, MoCo.V3~\cite{iccv2021mocov3} and DINO~\cite{iccv2021dino} are proposed to tackle unstable issues and degenerated performances of CL based on popular Vision Transformers~\cite{iclr2021vit}.

\vspace{-0.5em}
\paragraph{Mixup.} 
MixUp~\cite{zhang2017mixup}, convex interpolations of any two samples and their unique one-hot labels were presented as the first mixing-based data augmentation approach for regularising the training of networks. ManifoldMix~\cite{verma2019manifold} and PatchUp~\cite{faramarzi2020patchup} expand it to the hidden space. CutMix~\cite{yun2019cutmix} suggests a mixing strategy based on the patch of the image, \textit{i.e.}, randomly replacing a local rectangular section in images. Based on CutMix, ResizeMix~\cite{qin2020resizemix} inserts a whole image into a local rectangular area of another image after scaling down. FMix~\cite{harris2020fmix} converts the image to Fourier space (spectrum domain) to create binary masks. 
To generate more semantic virtual samples, offline optimization algorithms are introduced for the saliency regions. SaliencyMix~\cite{uddin2020saliencymix} obtains the saliency using a universal saliency detector. With optimization transportation, PuzzleMix~\cite{kim2020puzzle} and Co-Mixup~\cite{kim2021comixup} present more precise methods for finding appropriate mixup masks based on saliency statistics. SuperMix~\cite{dabouei2021supermix} combines mixup with knowledge distillation, which learns a pixel-wise sample mixing policy via a teacher-student framework. More recently, TransMix~\cite{2021transmix} and TokenMix~\cite{eccv2022tokenmix} are proposed specially designed Mixup augmentations for Vision Transformers~\cite{iclr2021vit}. Differing from previous methods, AutoMix~\cite{liu2022automix} can learn the mixup generation by a sub-network end-to-end, which generates mixed samples via feature maps and the mixing ratio.
Orthogonal to the sample mixing strategies, some researchers \cite{aaai2022grafting, 2021transmix, 2022decouplemix} improve the label mixing policies upon linear mixup.

\vspace{-0.5em}
\paragraph{Mixup for contrastive learning.}
A complementary method for better instance-level representation learning is to use mixup on CL~\cite{nips2020mochi, 2022aaaiunmix}. When used in collaboration with CE loss, Mixup and its several variants provide highly efficient data augmentation for SL by establishing a relationship between samples. Most approaches are limited to linear mixup methods without a ground-truth label. For example, Un-mix~\cite{2022aaaiunmix} attempts to use MixUp in the input space for self-supervised learning, whereas the developers of MoChi~\cite{nips2020mochi} propose mixing the negative sample in the embedding space to increase the number of hard negatives but at the expense of classification accuracy. i-Mix~\cite{2021iclrimix}, DACL~\cite{icml2021dacl}, BSIM~\cite{2020bsim} and SDMP~\cite{cvpr2022sdmp} demonstrated how to regularize contrastive learning by mixing instances in the input or latent spaces. We introduce an automatic mixup for SSL tasks, which adaptively learns the instance relationship based on inter- and intra-cluster properties online.



\end{document}